\documentclass[preprint,12pt,authoryear]{elsarticle}

\usepackage{tikz}
\usetikzlibrary{positioning, arrows.meta}
\usepackage[percent]{overpic}

\usepackage{tabularx} 
\usepackage{amssymb}
\usepackage{amsmath}
\usepackage{natbib}
\usepackage{bm}
\usepackage{booktabs}
\usepackage{multirow}
\usepackage{array}
\usepackage{threeparttable}
\usepackage{lineno}
\usepackage{soul}
\usepackage{color}
\usepackage[dvipsnames]{xcolor}
\sethlcolor{white}

\begin{document}


\begin{frontmatter}



\title{Realistic pedestrian-driver interaction modelling using multi-agent RL with human perceptual-motor constraints}

\author[inst1]{Yueyang Wang}
\author[inst2]{Mehmet Dogar}
\author[inst1]{Gustav Markkula}

\affiliation[inst1]{organization={Institute for Transport Studies},
            addressline={University of Leeds}, 
            city={Leeds},
            postcode={LS2 9JT}, 
            state={Leeds},
            country={UK}}

\affiliation[inst2]{organization={School of Computer Science},
            addressline={University of Leeds}, 
            city={Leeds},
            postcode={LS2 9JT}, 
            state={Leeds},
            country={UK}}


\begin{abstract}
Modelling pedestrian-driver interactions is critical for understanding human road user behaviour and developing safe autonomous vehicle systems. Existing approaches often rely on rule-based logic, game-theoretic models, or ‘black-box’ machine learning methods. However, these models typically lack flexibility or overlook the underlying mechanisms, such as sensory and motor constraints, which shape how pedestrians and drivers perceive and act in interactive scenarios. In this study, we propose a multi-agent reinforcement learning (RL) framework that integrates both visual and motor constraints of pedestrian and driver agents. Using a real-world dataset from an unsignalised pedestrian crossing, we evaluate four model variants—one without constraints, two with either motor or visual constraints, and one with both—across behavioural metrics of interaction realism. Results show that the combined model with both visual and motor constraints performs best. Motor constraints lead to smoother movements that resemble human speed adjustments during crossing interactions. The addition of visual constraints introduces perceptual uncertainty and field-of-view limitations, leading the agents to exhibit more cautious and variable behaviour, such as less abrupt deceleration. In this data-limited setting, our model outperforms a supervised behavioural cloning model, demonstrating that our approach can be effective without large training datasets. Finally, our framework accounts for individual differences by modelling parameters controlling the human constraints as population-level distributions, a perspective that has not been explored in previous work on pedestrian-vehicle interaction modelling. Overall, our work demonstrates that multi-agent RL with human constraints is a promising modelling approach for simulating realistic road user interactions.
\end{abstract}


\end{frontmatter}

\newpage
\section{Introduction}
\label{sec:Introduction}

\subsection{Background}
\label{subsec:Background}

The development of autonomous vehicles (AVs) has gained significant momentum in recent years, positioning them as a key component of future mobility systems. One major challenge in deploying AVs is the lack of realistic models that can capture the subtle and dynamic interactions between human road users in mixed traffic \citep{rasouli2019autonomous}. Pedestrians deserve particular attention because they are vulnerable road users and often behave less predictably than drivers, who are constrained by vehicle dynamics and traffic regulations \citep{gonzalez2021agent}. To develop realistic pedestrian models, it is therefore crucial to account for the mutual adaptations between pedestrians and drivers, which fundamentally shape crossing decisions and interaction patterns in real-world traffic.

Current approaches to modelling road user interaction can be broadly categorised into two types: mechanistic models and data-driven machine learning (ML) models. Mechanistic models include cognitive models, physics-based models, and game-theoretic models. These models rely on strong behavioural assumptions. For example, cognitive models are based on hypotheses about underlying mental processes and aim to represent how human behaviour is generated \citep{pekkanen2022variable,markkula2023explaining}. Such models offer high interpretability and can provide deeper insights into the mechanisms behind observed behaviours. However, the high complexity of these models limits their scalability and generalisability.

In contrast, data-driven models, such as LSTMs and Transformers, learn behavioural patterns directly from data \citep{yang2025interact,divya2025gaze}. Given sufficient data and computational resources, they can achieve high predictive accuracy on trajectory-based metrics such as root mean square error (RMSE). Nevertheless, they often lack interpretability, which is a key requirement in safety-critical applications like autonomous driving \citep{aradi2020survey,itkina2023interpretable}. In addition, their performance tends to degrade in data-limited or rare situations, which often correspond to safety-critical events where accurate behavioural prediction is important.

Another class of algorithms with the potential to bridge the aforementioned gap is reinforcement learning (RL). Modelling human behaviour using RL builds on the theoretical assumption that humans act to maximise expected rewards, while deep RL provides a powerful computational approach to approximate such reward-maximising behaviour policies \citep{sutton2018reinforcement,howes2023towards}. This framework offers the flexibility to generalise across a wider range of contexts and scenarios \citep{chen2015emergence,chen2021apparently,jokinen2021multitasking}.

In our previous work, we developed the pedestrian crossing model using RL that incorporated human perceptual and motor constraints, resulting in human-like crossing timing and speed \citep{wang2025pedestrian,wang2025Modeling}. However, these studies were limited in two important ways. First, only the pedestrian was modelled with a single-agent policy, while the vehicle followed a preprogrammed trajectory, preventing realistic mutual adaptation between the two agents. Second, they were evaluated using data from controlled experiments under simplified laboratory conditions, which limited the diversity of interaction contexts. Third, although between-individual variability was considered by fitting fixed non-policy parameters for each participant across repeated trials, this approach relied on multiple observations per individual and does not generalise well to real-world situations.

In this work, we address these limitations by modelling both pedestrian and vehicle behaviours within a multi-agent RL framework, where both agents adapt to each other’s actions using a real-world dataset of unsignalised crossing scenarios. Building on prior efforts to incorporate human perceptual and motor constraints, this study is the first to embed such constraints within a fully interactive, two-agent setting grounded in real-world data. Furthermore, we introduce a novel population-level parameter fitting procedure that captures between-individual variability by modelling human constraint-related parameters as distributions across the population, from which each individual’s parameters are drawn. This approach makes it possible to model individual differences in road user behaviour even when only a small amount of data is available.

\subsection{Related work}
\label{subsec:Related work}

Given the importance of pedestrian-vehicle interaction in the safe deployment of AVs, various modelling approaches have been developed to capture and predict such behaviour. In this section, we review relevant studies, organised by methodological category.

\subsubsection{Mechanistic models}
\label{subsec:Mechanistic models}

Mechanistic models build on physics, psychology, or cognitive science, and explain road user behaviour through explicit mathematical formulations. These approaches offer interpretability and explanation of the behaviour but often struggle to capture the flexibility of human behaviour in dynamic traffic environments.

Early models often used logistic regression to predict binary crossing decisions, typically based on variables such as vehicle speed and distance \citep{lobjois2007age,tian2022impacts}. Extensions of these models incorporated perceptual cues to explain why certain decisions are made \citep{tian2022explaining}. However, these models are inherently limited in scope, as their binary outputs and reliance on simple decision thresholds make them unsuitable for simulating the continuous and adaptive nature of real-world road behaviour.

Game-theoretic models, which simulate strategic reasoning between interacting agents, have also been applied. For example, \citet{fox2018should} modelled pedestrian-AV interactions to support safe vehicle planning. More recently, \citet{dang2025dynamic} proposed a dynamic game-theoretic framework incorporating bounded rationality, improving upon classical models by relaxing the assumption of fully rational agents. However, several practical limitations persist. For instance, the model assumed that agents have perfect sensory information and can execute actions without motor constraints. Moreover, pedestrian actions were restricted to a low-dimensional, discrete set of walking speeds, without the possibility to continuously adjust speed or direction, primarily due to the computational difficulty in solving game-theoretic formulations in high-dimensional continuous spaces.

Evidence accumulation models, grounded in psychology and cognitive science, have also been applied to road user behaviour modelling \citep{giles2019zebra,pekkanen2022variable}. In this approach, decision-making is modelled as a noisy integration of sensory and contextual information over time, terminating in an action once a decision threshold is reached. These models aim to capture the gradual nature of human decision processes and offer insight into the underlying cognitive mechanisms.
However, modelling complex road-user interactions using evidence accumulation alone remains challenging. To address this, \citet{markkula2023explaining} proposed an integrated model that combines noisy perception, evidence accumulation, and game-theoretic reasoning, enabling the reproduction of several empirically observed behaviours. The authors showed that to simulate even highly simplified pedestrian-vehicle interactions (e.g., requiring both agents to travel along straight perpendicular paths) requires the integration of multiple theoretical frameworks. Due to their structural complexity and reliance on strong assumptions, such mechanistic models are difficult to generalise and scale to more varied and complex real-world traffic environments.

\subsubsection{Data-driven models}
\label{subsec:Machine learning models}

In this paper, we define pure data-driven models as those that directly learn the mapping from input to output using ground-truth data, typically through neural networks trained on observed motion histories, without relying on handcrafted rules or explicit behavioural assumptions.

A common strategy for modelling road user interaction is to formulate trajectory prediction as a time-series forecasting problem, in which the model predicts future trajectories over a fixed horizon based on preceding motion. For example, \citet{deo2018convolutional} proposed an LSTM-based architecture that captures spatial relationships between vehicles to forecast human-driven vehicle trajectories. Other architectures have also been explored for interaction modelling. Graph-based neural networks represent the relational structure among agents in a scene \citep{yi2016pedestrian,ye2021gsan}, while more recently, Transformer-based models have gained attention for pedestrian behaviour modelling due to their ability to capture long-range dependencies and efficiently process sequential data \citep{yang2025interact,divya2025gaze}.

Although these models achieve strong performance on trajectory-based accuracy metrics, several limitations remain. First, data-driven models are often criticised as `black boxes': they lack interpretability and provide little insight into the underlying mechanisms of the behaviours they predict \citep{madala2023metrics}. This lack of transparency is concerning in the context of autonomous driving, since behaviour models are often used within AV decision-making pipelines, and unexplained prediction errors may propagate into AV planning and pose safety risks \citep{roshdi2024road}. Second, high predictive accuracy on an aggregate level does not necessarily imply that the model captures behaviourally meaningful features of an interaction \citep{srinivasan2023beyond}. For example, as shown by \citet{srinivasan2023beyond}, machine-learned trajectory predictors with low RMSE can still fail to reproduce human-like interaction patterns such as courtesy lane changes or collision aversion. Third, the absence of theoretical grounding can reduce the robustness of such models. Although they may perform well on training data, their generalisability to rare or unseen scenarios is often limited \citep{xu2020explainable}, and they depend heavily on large, diverse training datasets. In practice, collecting sufficient data to cover the full range of road user behaviours is challenging, as many safety-critical or rare events are underrepresented, or entirely missing, from existing datasets \citep{diaz2022ithaca365}.

\subsubsection{Reinforcement learning models}
\label{subsec:Reinforcement learning models}

Different from data-driven models, RL agents learn decision policies through trial-and-error interactions with the environment in order to maximise cumulative reward. Owing to its sequential decision-making nature and its ability to capture interactions between agents and their environment, RL has been applied to simulate road user interactions \citep{charalambous2019did,vizzari2022pedestrian}.

One major challenge in behaviour modelling with RL lies in designing reward functions that realistically reflect human intentions. Inverse reinforcement learning (IRL) addresses this by inferring latent reward functions from demonstrated behaviour \citep{ng2000algorithms}. For instance, \citet{nasernejad2021modeling,nasernejad2023multiagent} used IRL to model pedestrian-vehicle interactions in near-miss scenarios, recovering reward structures from real-world demonstrations. These models are typically formalised as Markov Decision Processes (MDPs), which assume full observability of the environment. However, human decision-making often occurs under uncertainty, making Partially Observable MDPs (POMDPs) a more appropriate framework. Consequently, by assuming full access to environmental information, MDP-based RL models overlook key human limitations such as perceptual uncertainty, leading to reward functions that may misrepresent the decision-making processes of real road users.

One line of work which has adopted the POMDP perspective and used RL to develop human models is computational rationality, which assumes that humans behave (near-) optimally given internal utility functions and cognitive or physical constraints \citep{gershman2015computational,oulasvirta2022computational}. This approach was initially applied in human-computer interaction \citep{chen2015emergence,tseng2015adaptation,jokinen2020adaptive}. More recently, we have applied it to pedestrian crossing decisions, incorporating cognitive mechanisms such as perceptual noise \citep{wang2023modeling,wang2025pedestrian}. We subsequently extended this approach to simulate the full pedestrian crossing process, including responses to lighting conditions and external human-machine interfaces \citep{wang2025Modeling}. However, these models were developed and validated in controlled experimental settings: Participants always started from standstill at the kerb, omitting the pedestrian approach phase; participants experienced multiple similar interactions during the experiment, which may have caused some learning effects. Moreover, the car was preprogrammed and did not respond to the pedestrian's behaviour, whereas two-sided interaction is known to be important for real-world driver-pedestrian interaction \citep{schneemann2016analyzing}.

Two-sided, or joint, behaviour is a key characteristic of natural traffic interactions, as pedestrians and drivers continuously adapt to each other’s actions \citep{domeyer2022driver}. This has motivated the use of multi-agent reinforcement learning (MARL) to model interactive road user behaviour. Recent MARL studies have explored cooperative and competitive interactions among vehicles or between pedestrians and vehicles \citep{cornelisse2024human,sackmann2024learning}, showing that multi-agent frameworks can capture the bidirectional adaptation that is absent in single-agent formulations. However, these approaches typically focus on optimising interaction outcomes or reproducing empirical patterns, without accounting for human perceptual or motor constraints that shape how such adaptations emerge. In contrast, past computational rationality models have focused on human perceptual and motor limitations but have rarely extended to fully interactive, two-agent settings. The present study bridges this gap by combining computational rationality with a MARL framework, enabling both pedestrian and driver agents to adapt to each other under human-like sensory and motor constraints.

\subsection{Contribution of this work}
\label{subsec:Contribution of this work}

This study addresses key limitations in existing models of pedestrian-vehicle interaction by formulating the problem as a multi-agent RL task, in which both the pedestrian and vehicle are modelled as agents subject to visual and motor constraints. We focus specifically on one-to-one interactions between a single pedestrian and a single vehicle. In addition, we introduce gaze-dependent acuity, moving beyond the simplifying assumption in our earlier controlled-experiment modelling that pedestrians continuously fixate on the vehicle. Unlike such prior assumptions, our model allows pedestrians to control their own gaze direction, making perceptual uncertainty dependent on both distance and gaze allocation. This represents a novel contribution that is essential for modelling realistic road user interaction, where pedestrians approach from different positions and are not always looking directly at the vehicle

To enable this modelling approach, we carefully selected and preprocessed a real-world trajectory dataset that enables the extraction of such one-to-one interaction events while minimising the influence of other road users and road layout. This allows us to construct a focused learning environment that aligns closely with the modelling assumptions and objectives.

Importantly, to the best of our knowledge, this is the first study to bring a bounded optimality perspective into a multi-agent RL framework for modelling human-human and human-machine interaction. By integrating cognitive constraints, real-world data, and multi-agent interaction, this work represents a meaningful step towards making the framework applicable to AV development, including both testing and optimisation.

\section{Methods}
\label{sec:Methods}

This section presents our modelling framework for pedestrian-vehicle interaction. 
Section~\ref{subsec:Overview of the method} provides a high-level overview of the framework. Section~\ref{subsec:Model assumptions} outlines the cognitive and motor constraints that shape the agents’ behaviour, and Section~\ref{subsec:Nonpolicy parameters} introduces the parameters specifying these human constraints. Section~\ref{subsec:Model variants} defines the model variants used for evaluation. Section~\ref{subsec:Dataset} then describes the real-world dataset and the extraction of interaction trajectories. The reinforcement learning problem and algorithm are detailed in Sections~\ref{subsec:Reinforcement learning problem} and~\ref{subsec:Reinforcement learning algorithm}, followed by parameter fitting (Section~\ref{subsec:Fitting of non-policy parameters}) and the behavioural cloning model (Section~\ref{subsec:Behavioural cloning model}).

\subsection{Overview of the method}
\label{subsec:Overview of the method}

\begin{figure}[!t]
      \centering
      \includegraphics[scale=0.45]{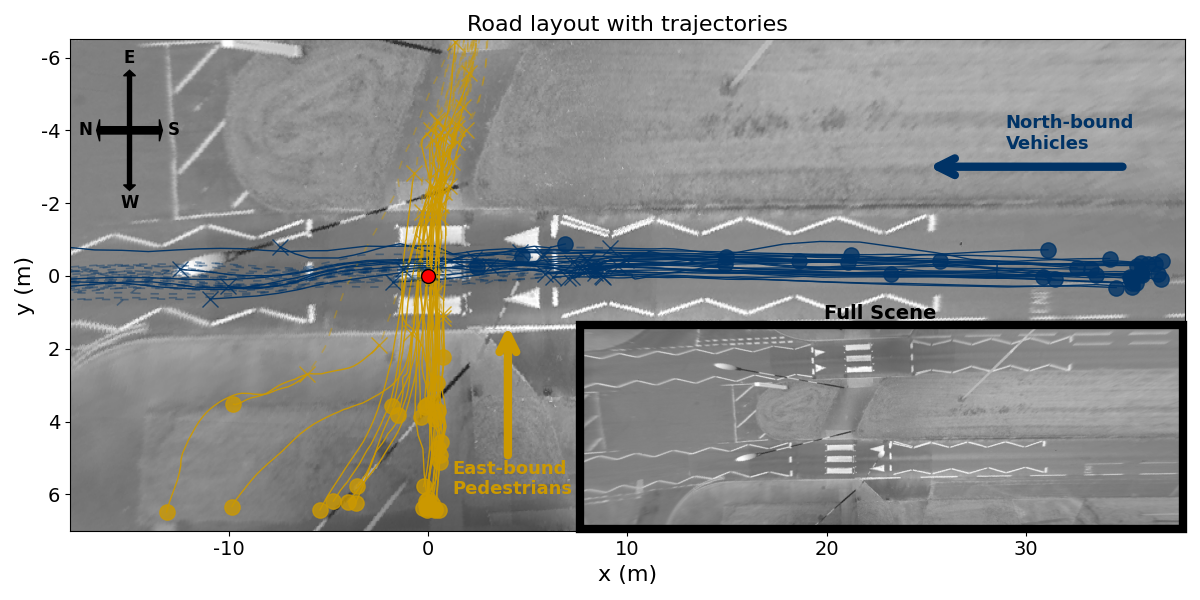}
      
      \caption{Real-world pedestrian (yellow) and vehicle (blue) trajectories overlaid on an image of the road layout at the study site. Solid lines represent trajectory segments within the first 6 seconds of the interaction, while dashed lines indicate segments beyond this window. The ‘o’ markers denote the starting points and the ‘×’ markers denote the end points of the 6-second segments. The red dot marks the centre of the zebra crossing, referred to in the text as the crossing point. A zoomed-out view of the full scene is shown in the bottom right for context. The compass rose at the top left indicates the cardinal directions.}
      \label{fig:Road_layout}
\end{figure}

We formulate pedestrian-vehicle interaction as a multi-agent reinforcement learning (RL) problem. The framework includes two policies: one for the pedestrian, $\pi_\mathrm{ped}$, and one for the vehicle, $\pi_\mathrm{veh}$. These policies are trained through trial-and-error interaction in a simulated environment that replicates the geometry of the real-world location from which we collected our dataset (as shown in Fig.~\ref{fig:Road_layout}), and the initial conditions in that dataset.

These take their simplest form in the No-Constraint (NC) variant of our model, where the pedestrian and the vehicle are represented as agents with ideal, non-constrained perception and motor execution. In this baseline model variant, the pedestrian observes its own position and speed together with the vehicle’s position and speed, and chooses continuous values of walking speed and direction. The vehicle observes its own position and speed and the pedestrian’s position and speed, and selects a continuous acceleration value. This NC formulation is straightforward but unrealistic, as it ignores many of the perceptual and motor constraints that shape real human behaviour.

To make the RL agent more human-like, we extended the NC framework with additional assumptions, leading to three constrained model variants: the Visual-Constraint (VC) model, the Motor-Constraint (MC) model, and the full Visual-and-Motor-Constraint (VMC) model, as further described below.

\subsection{Mechanistic assumptions of human constraints}
\label{subsec:Model assumptions}

This section outlines the key mechanistic assumptions in our model of pedestrian-vehicle interactions. These assumptions reflect perceptual, motor, and cognitive limitations derived from empirical and theoretical research, which shape how agents perceive the environment and select actions. Some of these mechanisms, such as noisy perception, ballistic speed control, and walking effort, were introduced in our prior work~\citep{wang2025pedestrian,wang2025Modeling}, while others, including gaze-dependent acuity and constraints on driver acceleration control, are new additions aimed at supporting more realistic two-agent interactions. The remainder of this section describes each of these mechanisms in detail.

\subsubsection*{Noisy visual input}Human perception is inherently uncertain and subject to visual noise \citep{faisal2008noise}. Prior studies have suggested that these perceptual limitations influence pedestrian crossing decisions \citep{kotseruba2023intend}. To capture this effect, we introduced angular noise at the level of the human retina \citep{kwon2015unifying}, which was also implemented in our previous pedestrian behaviour models \citep{wang2025pedestrian, wang2025Modeling}. Specifically, the perceived position of the other agent is derived from the visual angle below the horizon \citep{ooi2001distance, markkula2023explaining}, with a constant Gaussian angular noise of standard deviation, $\nu$, which could vary between agents. In practice, this means that the ego agent perceives the other agent’s position along its line of approach with a noise of standard deviation \citep{markkula2023explaining}:

\begin{equation}
\label{eq:noisy_perception}
\sigma_x(t) = |d_\mathrm{l}(t)| \left( 1-\frac{h}{d(t)\cdot\tan(\arctan\frac{h}{d(t)}+\nu)}\right),
\end{equation}
\noindent where $d_\mathrm{l}(t)$ is the longitudinal distance from the other agent to the crossing point (i.e., along the $y$-axis for the pedestrian and the $x$-axis for the vehicle in Fig.~\ref{fig:Road_layout}), $d(t)$ is the distance between two agents, and $h$ is the eye height over the ground of the ego agent, set to $1.6$~m for the pedestrian agent and $1.2$~m for the vehicle agent, assuming homogeneous heights for simplicity.

While the formulation above captures perception uncertainty, our earlier modelling based on controlled experimental data assumed that pedestrians continuously fixated on the approaching vehicle, which was reasonable since participants interacted with a single car and needed to monitor it to ensure safe crossing. In contrast, the present work focuses on real-world data, where pedestrians approach the crossing from different positions, sometimes such that the vehicle is initially even behind the pedestrian, and are therefore not necessarily fixating on the vehicle at all times. To account for this, we modelled the decline in visual acuity as a function of gaze eccentricity, making visual uncertainty dependent not only on distance but also on whether the vehicle falls within or outside the centre of the pedestrian's gaze. Human perception is most accurate at the fovea and becomes less precise with increasing distance from the centre of gaze. We adopted the midget retinal ganglion cell (RGC) density model proposed by \citet{watson2014formula}, which relates receptive field density \( d_{gf}(\epsilon) \) to the eccentricity angle \( \epsilon \) in degrees.

Following \citet{watson2014formula}, we treated visual acuity as proportional to the square root of RGC density. To normalise acuity to 1 at the fovea, we define: \(
\alpha(\epsilon) = \frac{\sqrt{d_{gf}(\epsilon)}}{\sqrt{d_{gf}(0)}}\), where \( d_{gf,0} \) is the peak density at zero eccentricity. The RGC density function is defined as:

\[
d_{gf}(\epsilon) = d_{gf,0} \cdot \left[a_k \cdot \left(1 + \frac{|\epsilon|}{r_{2,k}}\right)^{-2} + (1 - a_k) \cdot \exp\left(-\frac{|\epsilon|}{r_{e_k}}\right)\right]
\]

\noindent using parameters from \citet{watson2014formula}: \( a_k = 0.9729 \), \( r_{2,k} = 1.084 \), \( r_{e_k} = 7.633 \), and \(d_{gf,0}= 33163.2 \, \text{deg}^{-2} \). For modelling purposes, we approximate the perceptual noise level at a given eccentricity as the inverse of acuity: \(\text{noise}(\epsilon) = \frac{1}{\alpha(\epsilon)}\). This effect was incorporated by scaling the visual noise from Equation~\ref{eq:noisy_perception} by an acuity-based factor:

\begin{equation}
\label{eq:acuity}
\sigma_x^\text{mod}(t, \epsilon) = \sigma_x(t) \cdot \left( \frac{1}{\alpha(\epsilon)} + \delta \right)
\end{equation}

\noindent where \( \epsilon \) is the angular deviation between the pedestrian’s gaze direction and the vehicle, \(\alpha(\epsilon)\) is the relative visual acuity at angle \(\epsilon\) as defined above, and \(\delta\) is a very small constant (\(10^{-5}\)) added as a minor implementation safeguard to ensure numerical robustness during simulation; it is not theoretically required and does not affect the results. Note that for simplicity, we only used Equation~\ref{eq:acuity} for the pedestrian agent; for the driver agent we used Equation~\ref{eq:noisy_perception}.

\subsubsection*{Bayesian visual perception}Beyond the introduction of noise in the visual input, it is also important to consider how humans interpret such uncertain information. There is evidence that the human perception system interprets its noisy input in a Bayes-optimal manner, and Bayesian inference has been widely used to model perception and sensorimotor decision-making~\citep{kwon2015unifying,knill2004bayesian,stocker2006noise}. In line with our previous work on modelling pedestrian perception under uncertainty~\citep{wang2025Modeling, wang2025pedestrian}, we employed a Kalman filter to model visual perception~\citep{kwon2015unifying,markkula2023explaining}. At initialisation, perceived position and speed are drawn from Gaussian distributions centred at the true values, with standard deviations determined by the visual noise and by the variability of speed observed in the data, respectively. The Kalman filter then receives a new noisy position observation at each simulation time step, with noise as described above, and produces estimates of the other agent's position and velocity, along with their respective uncertainties. These filtered estimates represent the agent's belief state, serving as a Bayes-optimal inference about the environment under uncertainty.


\subsubsection*{Ballistic walking speed control}Human motor control is often characterised by intermittent, ballistic adjustments rather than continuous fine-tuning \citep{tustin1947nature,craik1948theory}. In line with this view, we assumed that each walking step is ballistic, such that once initiated, the pedestrian maintains a fixed acceleration throughout the step. While some studies suggest humans may adjust trajectories mid-step \citep{barton2019control}, this behaviour is omitted here for simplicity. In our model, after selecting an action, the pedestrian applies a fixed acceleration to reach the desired speed, calculated as: \( a_{\mathrm{ped,}t^*} = \frac{v_\mathrm{ped}(t^*) - v_\mathrm{ped}(t^*-1)}{T_{\text{step}}} \), where $v_\mathrm{ped}(t^*)$ and $v_\mathrm{ped}(t^*-1)$ represent the velocities of the agent at two consecutive decision points. Here, \(t^*\) indexes decision points to distinguish from the simulation time step \(t\) in the RL environment. \( T_\mathrm{step}\) is the duration of a walking step, which is derived from empirical relations between step length and speed, with step length \( s \) estimated by~\cite{grieve1968gait}:  \( s = {v_{\text{ped}}}^{0.42} \). This yields \(T_\mathrm{step} = v_\mathrm{ped}^{-0.58}\).

\subsubsection*{Walking effort}Human walking behaviour reflects a trade-off between travel time and energetic cost~\citep{carlisle2023optimization}. Prior research has incorporated walking effort into biomechanical models of gait~\citep{tae_ho_choi_1997,salman_faraji__2018}. In our model, we adopted the biomechanical framework from~\citet{carlisle2023optimization} to represent walking effort as a cost associated with speed change during a step. The equation for the new velocity $v_i^{+}$ is: 
\(
{v_i^{+} = v_i^{-} \cos 2\theta + \sqrt{2u_i} \sin 2\theta}
\), where \( v_i^{-} \) is the initial walking speed, \( v_i^{+} \) is the new walking speed after the action, and 2\( \theta \) is the angle between two legs. The term \(\sqrt{2u_i}\) represents the speed change due to exerted effort. This formulation stems from modelling the leg as a pendulum, where effort \( u_i \) corresponds to the kinetic energy required to alter walking speed: 
\(
u_i = \frac{1}{2} m (\Delta v)^2
\)
and solving for \( \Delta v \):
\( \Delta v = \sqrt{2u_i / m} \). If we instead express effort per unit mass, we define: 
\(
U_i = u_i/m,
\)
which simplifies to:
\( \Delta v = \sqrt{2U_i} \). Therefore, the effort \( U_i \) required for this change is given by:

\begin{equation}
\label{eq: kinetic energy 2}
U_i = \frac{{(v^{-} \cos(2\theta) - v^{+})}^2}{2 \cdot \sin^2(2\theta)}
\end{equation}
The total walking effort is defined as: \(E_w = w_\mathrm{ped} \cdot U_i\), where $w_\mathrm{ped}$ is a parameter that scales the walking effort for different individuals.  A higher value of $w_\mathrm{ped}$ reflects a stronger preference for conserving energy during crossing.

\subsubsection*{Acceleration control of the driver}Human driving behaviour is subject to perceptual, cognitive, and motor delays, meaning that, in contrast with most RL models of driving \citep{sackmann2024learning,cornelisse2024human}, control actions such as steering or acceleration are not applied instantaneously. In the broader driver modelling literature, these delays are often modelled using a low-pass filter to reflect gradual motor execution~\citep{sharp2005driver}. Following this approach, we modelled vehicle acceleration as a smoothed response toward a target acceleration value. At each time step, the driver selects a target acceleration $a_{\mathrm{target}}$ through its policy, and the actual acceleration $a_{\mathrm{veh,}t}$ evolves incrementally as:
\begin{equation}
a_{\mathrm{veh,}t} = a_{\mathrm{veh,}t-1} + \frac{a_{\mathrm{target}} - a_{\mathrm{veh,}t-1}}{w_{\text{veh}}},
\end{equation}
where $w_{\text{veh}}$ is the motor non-policy parameter representing the vehicle’s responsiveness. This formulation assumes that the driver cannot instantaneously achieve their intended acceleration, but must gradually adjust due to both human neuromuscular limitations and vehicle dynamics. The resulting $a_{\mathrm{veh,}t}$ is used to update the vehicle’s speed and position at each time step.

\subsection{Non-policy parameters of human constraints}
\label{subsec:Nonpolicy parameters}

We refer to the parameters that specify the human constraints as non-policy parameters, to distinguish them from the trainable weights and biases of the policy network. We considered four such parameters: $\nu_{\mathrm{ped}}$ and $\nu_{\mathrm{veh}}$, which are \emph{visual non-policy parameters} related to noisy visual input, and $w_{\mathrm{ped}}$ and $w_{\mathrm{veh}}$, which are \emph{motor non-policy parameters} that control motor execution effects, i.e., walking effort scaling for the pedestrian and acceleration smoothing for the vehicle.

In our previous work, we conditioned each agent’s RL policy on non-policy parameters that captured its perceptual and motor constraints \citep{wang2025pedestrian,wang2025Modeling}. These parameters were provided as additional inputs to the policy network during training, allowing the learned policy to adapt its behaviour according to different combinations of perceptual and motor limitations. After training, we optimised the non-policy parameter values for each participant in the dataset, by minimising the difference between simulated and observed trajectories, thereby obtaining per-individual estimates of perceptual and motor constraints.

\begin{figure}[!t]
    \centering
    \includegraphics[width=0.85\textwidth]{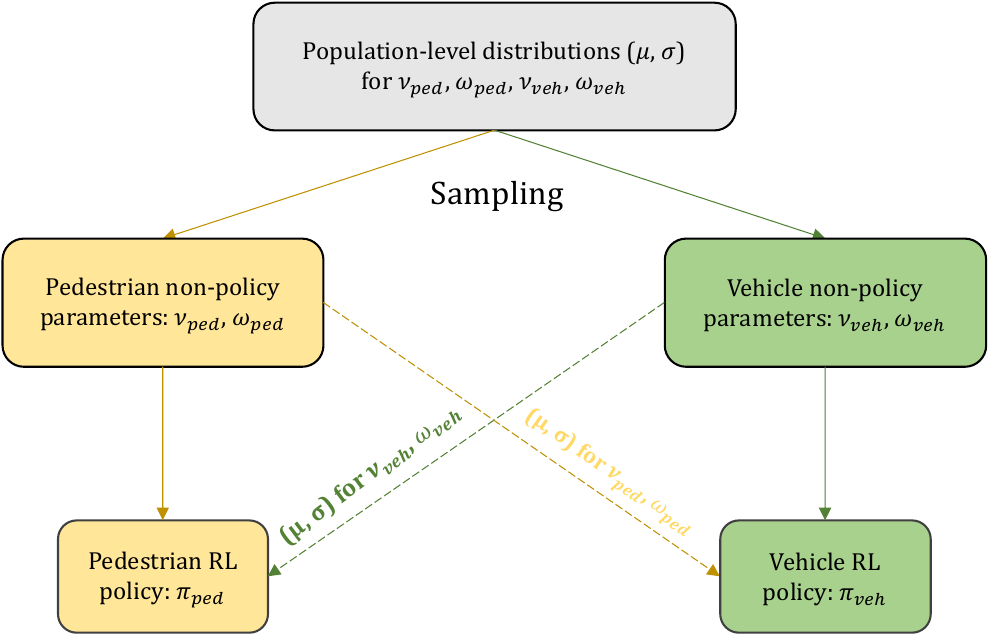}
    \caption{Overview of the two-stage sampling and conditioning scheme for non-policy parameters. Population-level means and standard deviations $(\mu, \sigma)$ are defined for $(\nu_\mathrm{ped}, w_\mathrm{ped}, \nu_\mathrm{veh}, w_\mathrm{veh})$. At the start of each episode, agent-specific values are sampled from these distributions. Each RL policy ($\pi_\mathrm{ped}$, $\pi_\mathrm{veh}$) is conditioned on its own parameters and on the population-level $(\mu, \sigma)$ of the other agent. Each observation $o$ also includes this information for conditioning.}
    \label{fig:Nonpolicy_scheme}
\end{figure}

In this work,  our dataset includes only a single observed interaction for each human agent, which does not permit us to fit per-individual parameter values. Instead, we directly model the population variability of the four non-policy parameters, as four separate normal distributions, and estimate the mean and standard deviation of these distributions from our data. 
Therefore, during RL policy training, at the start of each simulated interaction episode, we first sampled random values for these means and standard deviations. Specifically, the \emph{visual} non-policy parameters 
($\nu_\mathrm{ped}$, $\nu_\mathrm{veh}$) use means sampled uniformly from 
$[0.01,\ 0.1]$ and standard deviations from $[0.001,\ 0.01]$ (in radians). 
The \emph{motor} non-policy parameters are dimensionless: 
for the vehicle ($w_\mathrm{veh}$), means are sampled from $[1,\ 10]$ and 
standard deviations from $[0.1,\ 1]$; 
for the pedestrian ($w_\mathrm{ped}$), means from $[0.05,\ 0.5]$ and 
standard deviations from $[0.005,\ 0.05]$. Thereafter, individual parameter values were drawn from the resulting normal distributions and assigned to the agents in that episode (see Figure~\ref{fig:Nonpolicy_scheme} for an overview of this sampling and conditioning process). As in our previous work, each agent's policy is conditioned on its own non-policy parameter values, but here we also condition the policy on the population mean and standard deviation of the other agent's non-policy parameters. This reflects our assumption that each road user behaves in a way that is boundedly optimal given its own constraints, and given the distribution of constraints in the population of other road users.

\subsection{Model variants}
\label{subsec:Model variants}

\begin{figure}[!t]
      \centering
      \includegraphics[scale=0.55]{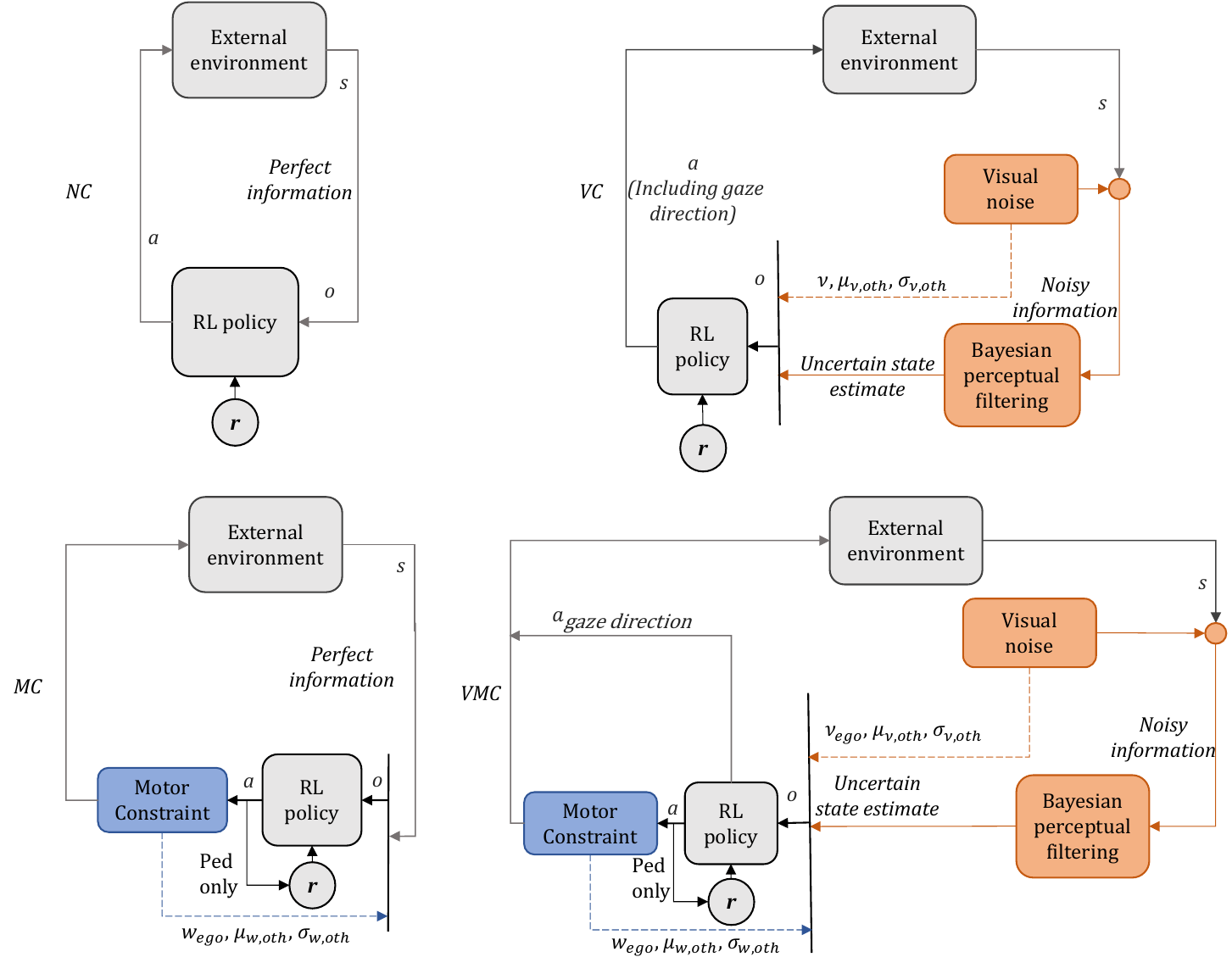}
      \caption{Comparison of model variants. Arrows follow the reinforcement learning loop: $s$ is the true environment state, $a$ is the agent action (including gaze direction in the VC and VMC models), $r$ is the received reward, and $o$ is the observation fed to the RL policy. In the MC and VMC models, the action $a$ contributes to the reward only for the pedestrian agent (walking effort penalty), as indicated by the `Ped only' label.}
      \label{fig:Model}
\end{figure}

To investigate how our different model assumptions influence the interaction between the pedestrian and vehicle, we categorised the assumptions into two types: \textbf{visual constraints} and \textbf{motor constraints}. The visual constraints include \textit{noisy visual input}, \textit{Bayesian visual perception}, while the motor constraints include \textit{walking effort}, \textit{pedestrian ballistic speed control}, and \textit{driver acceleration control}. Comparison of different model variants is shown in \figurename~\ref{fig:Model}. 

Based on the inclusion or exclusion of these constraints, we defined the following model variants:

\begin{itemize}
    \item \textbf{No-Constraint (NC) model}: A baseline model without visual or motor constraints.
    \item \textbf{Motor-Constraint (MC) model}: A model incorporating motor constraints only.
    \item \textbf{Visual-Constraint (VC) model}: A model incorporating visual constraints only.
    \item \textbf{Visual and Motor-Constraint (VMC) model}: A full model incorporating both visual and motor constraints.
\end{itemize}

\subsection{Dataset}
\label{subsec:Dataset}
This study focuses on modelling the interaction between one pedestrian and one vehicle. We used the dataset reported by \cite{kalantari2025testing}, which fits our aim of extracting one-one pedestrian-vehicle interaction with minimal effects from other road users and road layout. To obtain clean and representative pedestrian-vehicle interactions from real-world data, we employed a multi-stage trajectory extraction and filtering pipeline.

\figurename~\ref{fig:Road_layout} shows a top-down view of the study site in Leeds, UK, from which our data were collected. As shown in the bottom-right inset of \figurename~\ref{fig:Road_layout}, the site features a staggered pedestrian crossing with two single-lane zebra crossings. In this work, we only considered interactions at the \textit{Western} crossing, between North-bound vehicles (note that the UK has left-hand traffic) and East-bound pedestrians. The reasons for only including these interactions were the South-bound vehicles were also influenced by turning traffic
from the side street, which introduces variability in vehicle speed when approaching
the zebra crossing.

The dataset includes pedestrians crossing in both directions. Since pedestrians with different walking directions correspond to different origins and destinations, modelling both would require training two separate pedestrian policies. We excluded West-bound (from up to bottom in \figurename~\ref{fig:Road_layout}) pedestrians because they have already crossed another road and passed through the central refuge island, which exposes them to additional influences from South-bound vehicles. To maintain a clear focus on one-to-one interactions, we retained only interactions where the pedestrian walks from bottom to top in \figurename~\ref{fig:Road_layout} (labelled as East-bound in the dataset). In the original data, 311 East-bound pedestrian trajectories and 49,167 North-bound vehicles remained. To ensure that the retained pedestrian-vehicle pairs were not influenced by pedestrians walking in the opposite direction, we removed any North-bound vehicle approaching from the right (within 25 metres before the zebra crossing) if it temporally overlapped with at least one West-bound pedestrian in the central refuge island. 

For each pedestrian, we identified potential interacting vehicles based on their spatial proximity and temporal overlap. Specifically, a vehicle was considered part of the interaction if it was located within 25 m upstream of the zebra crossing, and a pedestrian was considered if they were within 20 m of the crossing along the $x$-axis and within 5 m of the kerb along the $y$-axis. Both agents also had to be present in the scene during an overlapping time interval. An interaction pair was retained only if exactly one vehicle satisfied these conditions for a given pedestrian. A second filtering step was then applied to ensure that each vehicle was matched with at most one pedestrian. Interactions involving multiple potential matches were excluded to maintain strict one-to-one interaction. After this step, 77 interaction pairs were left (each interaction pair consists of one East-bound pedestrian and one North-bound vehicle).

Following the above filtering steps, we further truncated each pedestrian trajectory to include only the portion where the pedestrian was within 10 metres of the crossing point, measured along the $x$-axis in \figurename~\ref{fig:Road_layout}. This ensured that the retained segments contained only the parts of the trajectories relevant to the crossing interaction. Each pedestrian-vehicle pair was then temporally aligned by retaining only the overlapping portion of their trajectories within the interaction zone. For each agent, the entry and exit times were defined as the moments of entering and leaving that zone, respectively, and the overlap was defined from the later entry to the earlier exit. From this synchronised window, we extracted the first 6 seconds, measured from the synchronised start time, in order to capture the early phase of interaction when both agents were simultaneously present. Interaction pairs with a total overlap shorter than 6 seconds were discarded to ensure sufficient behavioural content for modelling. This procedure left 34 interaction pairs, which are visualised in \figurename~\ref{fig:Road_layout}.

\subsection{Reinforcement learning problem}
\label{subsec:Reinforcement learning problem}

Reinforcement learning (RL) provides a computational framework in which an agent interacts with an environment by selecting actions to maximise cumulative rewards. In cases such as ours, where the agent does not have full information about its environment, the problem is typically formulated as a partially observable Markov decision process (POMDP), defined by a tuple $(\mathcal{S}, \mathcal{A}, \mathcal{T}, \mathcal{R}, \mathcal{O})$, where:

\subsubsection{State space ($\mathcal{S}$):}
\label{subsubsec:State space}

At each time step $t$, the environment state $s_t$ includes the true positions and velocities of both the pedestrian and the vehicle, as well as the simulation time. Specifically, the state comprises the pedestrian's position $(x_\mathrm{ped}, y_\mathrm{ped})$ and walking speed $v_\mathrm{ped}$, the vehicle's position $(x_\mathrm{veh}, y_\mathrm{veh})$ and speed $v_\mathrm{veh}$, the yaw angle of the pedestrian's walking direction $\theta_\mathrm{ped}$, and the vehicle's acceleration $a_\mathrm{veh}$. The walking angle $\theta_\mathrm{ped}$ is measured relative to the upward $y$-axis of the map (see \figurename~\ref{fig:Road_layout}), with $0^\circ$ indicating motion along the positive $y$-direction.

During training, initial conditions were sampled from a joint distribution fitted to the initial states of the pedestrians and vehicles at the start of each interaction segment in the real-world dataset. This distribution was fitted using kernel density estimation (KDE), covering start positions and speeds for both agents. This approach allowed the model to train on a diverse set of realistic interaction scenarios. During model testing, each simulation was initialised directly from the observed start state of the corresponding real-world interaction.

\subsubsection{Action space ($\mathcal{A}$):} 
\label{subsubsec:Action space}
The pedestrian's action $a_\mathrm{ped}$ consists of forward walking speed $v$ and a change in walking angle $\Delta\theta$, and optionally a gaze angle relative to the body angle $\phi$ (included in visual-constrained models, such as VC and VMC). Gaze orientation was not associated with any biomechanical effort or explicit cost, meaning the agent could freely direct its gaze. The driver's action $a_{\mathrm{veh,}target}$ is the target acceleration. It should be noted that in models with motor constraints (MC and VMC), the pedestrian speed action is not applied instantaneously. Instead, the selected speed only takes effect at the onset of a new walking step and remains constant during the execution of that step, consistent with the ballistic control assumption described in Section~\ref{subsec:Model assumptions}. This reflects the view that pedestrians cannot adjust their velocity continuously within a step but only between successive steps.

\subsubsection{Transition function ($\mathcal{T}$):} 
\label{subsubsec:Transition space}
The transition function defines how the current state \( s_t \) evolves into the next state \( s_{t+1} \) given the agents' actions \( a_t \).

In the NC and VC models, actions affect agent speeds and positions directly. In the MC and VMC models, pedestrian transitions are governed by ballistic execution: the walking speed is updated using a fixed acceleration over the step duration as explained above, and the walking angle is updated according to the selected change in walking angle. For the vehicle, the speed is updated using a smoothed acceleration \( a_{\text{veh}, t} \), as described in Section~\ref{subsec:Model assumptions}.

The simulation terminates when both agents reach their respective goal regions or a collision occurs. The pedestrian’s goal region is defined as having crossed the driving lane, while the vehicle’s goal region is defined as having passed beyond the zebra crossing.

\subsubsection{Reward function ($\mathcal{R}$):}

The reward function is designed to remain as simple as possible while encouraging successful and safe crossing and discouraging undesirable behaviours such as collisions, and rule violations. Most components build upon those used in our previous study \citep{wang2025Modeling}, which produced realistic and interpretable crossing behaviour. The present formulation retains these components while introducing additional terms to accommodate new action definitions, such as directional walking choices. To reflect the influence of time pressure on road-user decisions \citep{tian2022explaining,kong2023depth}, a time cost is included in the arrival rewards for both agents, making delayed arrivals less rewarding.

\paragraph{Pedestrian agent}
The pedestrian receives a positive reward upon successfully reaching the sidewalk on the opposite side:
\(r_\mathrm{arrive}^\mathrm{ped} = 40 - 0.5 \cdot t,\)
where \( t \) denotes the simulation time. The constant value of 40 was chosen based on initial tests with the NC model, which showed that this setting led to reasonable crossing behaviour. A penalty of \(r_\mathrm{collision} = -40\) is applied in the event of a collision with the vehicle. In addition, the pedestrian is rewarded for making forward progress before reaching the sidewalk on the opposite side:
\(
r_\mathrm{move} = \Delta y,
\)
where \( \Delta y \) is the change in longitudinal (up $y$-direction in \figurename~\ref{fig:Road_layout}) position over the previous time step. Unlike in the previous models \citep{wang2025Modeling}, this term is included here because the pedestrian can choose its walking direction. The forward progress reward therefore guides the pedestrian agent to cross the road in the desired direction rather than wandering laterally.

In models with motor constraints (MC and VMC), as previously described, a \emph{walking effort penalty} is introduced to represent the physical cost of accelerating from one walking speed to another \citep{carlisle2023optimization,wang2025Modeling}. At each action update, the ballistic model computes the required speed change, and the energy cost is penalised as: \(r_\mathrm{walk} = - E_w = - w_\mathrm{ped} \cdot U_i,\)
where \( w_\mathrm{ped} \) is the motor effort parameter, and \( U_i \) is the walking effort work calculated as described in Section \ref{subsec:Model assumptions}. This penalty is only applied at the start of each step. An additional penalty is applied for stepping onto the road outside the crosswalk at each time step: \(r_\mathrm{off} = -0.2\).

Summing over the terms, the total pedestrian reward is given by:
\begin{equation}
r_\mathrm{ped} = r_\mathrm{arrive}^\mathrm{ped} + r_\mathrm{move} + r_\mathrm{walk} + r_\mathrm{off} + r_\mathrm{collision}
\end{equation}

\paragraph{Vehicle agent}
Like the pedestrian, the vehicle receives a positive reward upon successfully crossing the designated zone:
\(
r_\mathrm{arrive}^\mathrm{veh} = 40 - 0.5 \cdot t.
\)
A collision incurs a penalty of $-40$. To represent the traffic rule requiring drivers to yield to pedestrians at zebra crossings, a penalty of \( r_\mathrm{nonyield} = -30 \) is imposed when the vehicle enters the crossing area without yielding to a nearby pedestrian. Specifically, this penalty is triggered once at the time step when the vehicle first enters a zone extending \( 5\,\mathrm{m} \) from the zebra crossing, provided that the pedestrian is located within \( 2\,\mathrm{m} \) in $x$ direction of the zebra crossing and within \( 3\,\mathrm{m} \) of the curb in the $y$ direction.

The total vehicle reward is:
\begin{equation}
r_\mathrm{veh} = r_\mathrm{arrive}^\mathrm{veh} + r_\mathrm{nonyield} + r_\mathrm{collision}
\end{equation}

\subsubsection{Observation space ($\mathcal{O}$):} 
\label{subsubsec:Observation space}

At each time step, both agents receive a normalised observation vector comprising (i) their own state, (ii) an estimate of the other agent’s state, and (iii) relevant non-policy parameters. Each observation also includes the current time step $t$, allowing the agent to account for the passage of time given that the reward function incorporates a time cost. All features are normalised to the range $[0,1]$ by linearly scaling each variable between its approximate minimum and maximum values, ensuring that negative values are also mapped within this range.

The \textbf{pedestrian} observes its own $x$ and $y$ coordinates, forward walking speed, walking direction, gaze yaw angle, and the vehicle’s $y$ coordinate and speed. In models without visual constraints, the vehicle state is observed without noise, whereas in models with visual constraints they are estimated from noisy input as described in Section~\ref{subsec:Model assumptions}. In the latter case, the observation additionally includes uncertainty measures from the Kalman filter, specifically the estimated variances of the vehicle’s position and speed. Execution-related features capture the pedestrian’s step cycle: in models with motor constraints, the observation provides the remaining execution time of the current walking step, indicating the time remaining until a new speed command can take effect. The observation also includes the pedestrian’s own non-policy parameters, as well as the mean and standard deviation of the vehicle’s non-policy parameter distributions, as explained in Section~\ref{subsec:Nonpolicy parameters}.

The \textbf{vehicle} observes its own $x$ coordinate, speed, acceleration, and, in models with motor constraints, also the target acceleration. In models with visual constraints, the vehicle does not observe the pedestrian state directly but instead receives Kalman-filtered estimates of the pedestrian’s $x$ and $y$ coordinates and speeds. Along with these estimates, the observation also includes the corresponding uncertainty measures from the Kalman filter, namely the estimated variances of the pedestrian’s position and speed. The pedestrian’s walking direction is also provided. The observation further includes the vehicle’s non-policy parameters, along with the mean and standard deviation of the pedestrian’s non-policy parameter distribution.

\begin{figure}[!t]
    \vspace{-0.2cm}
      \centering
      \includegraphics[scale=0.45]{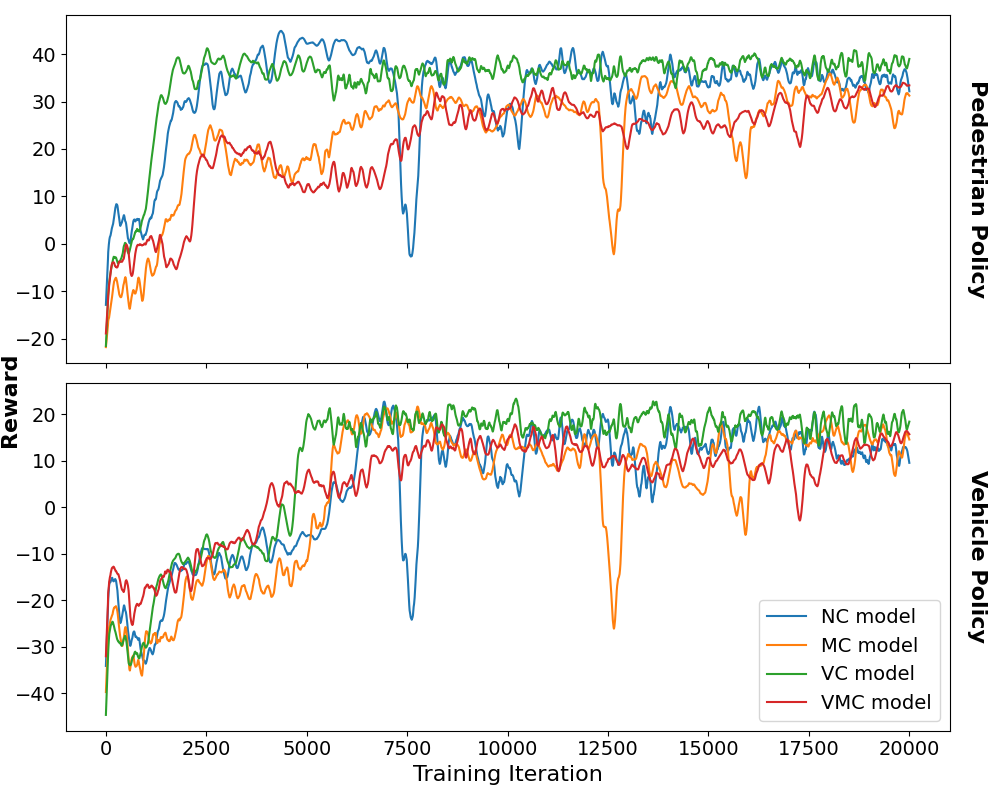}
       \vspace{-0.3cm}
      \caption{Reward plot.}
      \label{fig:Reward}
      \vspace{-0.2cm}
\end{figure}

\begin{table}[ht]
\vspace{-0.4cm}
\centering
\caption{Hyperparameters for the RL algorithm.}
\small
\begin{tabularx}{\textwidth}{XX}
\toprule
\textbf{Parameter} & \textbf{Value} \\
\midrule 
Algorithm & Soft Actor-Critic (SAC) \\
Training iterations & 20,000 \\
Hidden layers & [256, 128, 64] \\
Activation & ReLU \\
Learning rate & 0.001 \\
Discount factor ($\gamma$) & 0.995 \\
Batch size & 8192 \\
\bottomrule
\end{tabularx}
\label{tab:Hyperparameters}
\vspace{-0.4cm}
\end{table}

\subsection{Reinforcement learning algorithm}
\label{subsec:Reinforcement learning algorithm}

We used the Soft Actor-Critic (SAC) algorithm as implemented in Ray RLlib for all RL experiments \citep{liang2018rllib}. SAC optimises a stochastic policy in continuous action spaces and supports efficient, stable learning through off-policy updates \citep{haarnoja2018soft}. The reuse of past experiences may offer some benefits in our multi-agent setting, where interactions are complex \citep{christianos2020shared}. In our case, two policies were trained simultaneously, one for the pedestrian agent and one for the vehicle agent.

We trained the SAC model for 20,000 iterations. The reward curves of different model variants during training is shown in \figurename~\ref{fig:Reward}, illustrating the convergence trends of the different model variants. The hyperparameters used in our implementation are summarised in Table~\ref{tab:Hyperparameters}.

\subsection{Fitting of non-policy parameters}
\label{subsec:Fitting of non-policy parameters}
The values for the non-policy parameters that best describe the population are not learned through the RL training but are instead fitted by comparing model trajectories of the trained RL policy with real-world trajectory data.

To quantify how well the model behaviour matches real-world data, we adopted a composite Negative Log-Likelihood (NLL) metric inspired by recent work on evaluation of the performance of simulation-based agents \citet{montali2023waymo}. In this approach, comparison between model and real-world trajectories is conducted over a diverse set of behavioural metrics encompassing both motion dynamics and interaction outcomes, such as linear speed, acceleration, angular velocity, angular acceleration, distance to nearest object, time-to-collision (TTC), distance to road edge, offroad indication, and collisions. In \citet{montali2023waymo}, each of these metrics is first converted into a likelihood score by comparing simulated rollouts with logged data using kernel density estimates. NLLs are then averaged across time and agents, and finally aggregated into a single composite score via a weighted average. Safety-critical metrics such as collisions and offroad departures are assigned higher weights in their formulation.  

Building on this framework, we used the same likelihood-based approach and adapted the set of metrics to our specific pedestrian-vehicle interaction scenario. For simplicity, we did not introduce different weights for different metrics, effectively setting all weights to one. Following \citet{montali2023waymo}, we included metrics describing motion dynamics and interactions, namely the linear speed and linear acceleration of both agents, the pedestrian’s angular velocity and angular acceleration, and the inter-agent distance. In addition, we added a few metrics to better capture behaviours relevant to our task, including the $x$ positions of both agents, the pedestrian’s $y$ position and heading angle, and the projected Post-Encroachment Time (PET) \citep{lin2024real}. The projected PET is defined as the predicted time difference, based on current trajectories, between the pedestrian passing the $y$ coordinate of the crossing point and the vehicle passing its $x$ coordinate (the red dot in \figurename~\ref{fig:Road_layout}). A positive PET indicates that the pedestrian crosses first, while a negative PET indicates that the vehicle crosses first.

Given the scoring method described above, model fitting amounts to running simulations for different values of a non-policy parameter vector~$\boldsymbol{\phi}$ and finding the one that minimises the composite NLL. 
The vector 
\[
\boldsymbol{\phi} = [\mu_{\nu,{\mathrm{ped}}}, \sigma_{\nu,{\mathrm{ped}}}, \mu_{\nu,{\mathrm{veh}}}, \sigma_{\nu,{\mathrm{veh}}}, \mu_{w,{\mathrm{ped}}}, \sigma_{w,{\mathrm{ped}}}, \mu_{w,{\mathrm{veh}}}, \sigma_{w,{\mathrm{veh}}}],
\]
defines the means and standard deviations of the population distributions for the four non-policy parameters $(\nu_{\mathrm{ped}}, \nu_{\mathrm{veh}}, w_{\mathrm{ped}}, w_{\mathrm{veh}})$, giving eight free parameters in total.

Since the model contains stochasticity arising from both the noisy visual input and the trained SAC policy, we ran 5 rollouts for each parameter vector to obtain stable likelihood estimates. To mitigate the compounding divergence between model and real-world trajectories, model rollouts during testing were limited to 2 s in duration. Each 6 s real-world trajectory was therefore divided into three 2 s segments, with initial conditions taken from the real-world data and shared non-policy parameter values used across all three segments.

For each behavioural metric, a KDE was fitted to the real-world data using Scott’s rule for bandwidth selection \citep{scott2015multivariate}. The model distributions were then evaluated under the fitted KDEs to obtain per-feature likelihoods. The overall composite NLL was computed by averaging the NLL across all behavioural features and trajectory segments.

To optimise the non-policy parameter vector~$\boldsymbol{\phi}$, we used Bayesian optimisation with Gaussian processes via \verb|gp_minimize| from \texttt{scikit-optimize}. During optimisation, each element of~$\boldsymbol{\phi}$ was constrained within the ranges specified in Section~\ref{subsec:Nonpolicy parameters}. The optimisation was run for 500 iterations, with 100 initial random samples, using Expected Improvement as the acquisition function. The optimisation returns point estimates of these eight values, which define the population-level distributions from which individual agents sample their non-policy parameters during simulation. In addition to the composite NLL used as the optimisation objective, we also reported per-metric NLLs and the Kolmogorov-Smirnov (KS) statistic as complementary distributional measures, since KS is a widely used method for comparing empirical distributions and may be more familiar to some readers.

\subsection{Behavioural cloning model}
\label{subsec:Behavioural cloning model}
Behavioural cloning (BC) \citep{bain1995framework} was implemented as a supervised learning baseline for comparison with our RL framework. Its purpose was to evaluate to what extent a non-interactive imitation learning model, trained purely on the real-world demonstrations in our small dataset, could replicate realistic pedestrian and vehicle behaviours. Given the data limited setting, we trained and evaluated the BC baseline on the entire dataset to keep the training and evaluation conditions consistent with the RL models. This choice maximised access to available data and was expected to yield strong one-step (open-loop) predictions, whereas our main comparisons focus on closed-loop rollout performance on the same interaction set.

The BC models for both pedestrian and vehicle agents were trained separately, using the same data as for our RL model, but instead using supervised learning to predict the next agent action given the current state. The input features included both self and other-agent states: $x$ and $y$ position, speed, heading angle for the pedestrian; and $x$ position and speed for the vehicle, as well as time. These features correspond to the observation space of the NC variant of our RL model. The BC models produced the same action outputs as the NC model. Specifically, the pedestrian model predicted walking speed and heading angle, while the vehicle model predicted linear acceleration.

\begin{figure}[!t]
      \centering
      \includegraphics[scale=0.4]{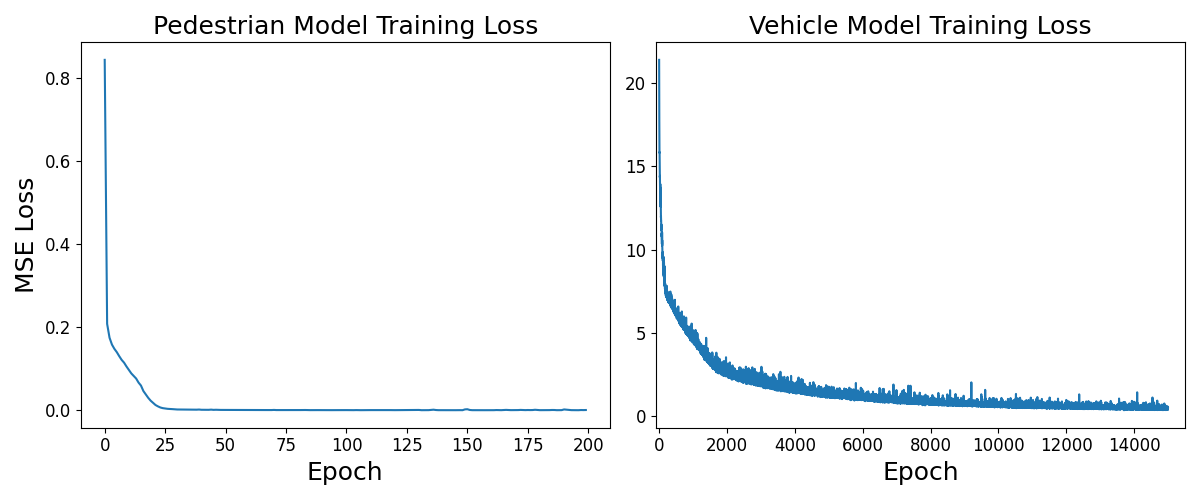}
       \vspace{-0.3cm}
      \caption{Training loss curves for the BC model.}
      \label{fig:loss}
      \vspace{-0.2cm}
\end{figure}

The pedestrian and vehicle BC models shared the same network architecture: two hidden layers with 64 units each and ReLU activation. They were trained using mean squared error (MSE) loss and the Adam optimiser with a learning rate of 0.001, until subjectively judged convergence, reached after 200 epochs and 15{,}000 epochs for the pedestrian and vehicle models, respectively. Training losses are shown in \figurename~\ref{fig:loss}.

\section{Results}
\label{sec:Results}

The results presented in this section are based on simulations using the trained RL model, with the fitted non-policy parameter distributions. For each real-world interaction, we initialised the simulation with the corresponding human starting conditions and ran the model five times, the same as in the fitting procedure, to account for variability arising from the visual noise and from the RL policy. The evaluation was performed on full interaction trajectories, without segmentation.

The same procedure was applied to the BC model, which is deterministic in its action predictions. In this case, the five runs per interaction yielded identical trajectories, but were repeated for consistency with the evaluation protocol used for the RL model.

\begin{table}[ht]
\vspace{-0.4cm}
\centering
\caption{Quantitative comparison of model performance across three behavioural metrics: negative log-likelihood (NLL), average displacement error (ADE), and final displacement error (FDE).}
\begin{tabularx}{\textwidth}{p{3.5cm}XXXXX}
\toprule
\textbf{Metric} & \textbf{BC} & \textbf{NC} & \textbf{VC} & \textbf{MC} & \textbf{VMC} \\
\midrule
NLL & 3.92 & 3.57 & 3.28 & 3.19 & \textbf{2.42} \\
ADE (m) & 5.92 & 5.17 & 4.60 & 5.61 & \textbf{2.87} \\
FDE (m) & 11.54 & 11.79 & 11.36 & 9.41 & \textbf{5.41} \\
\bottomrule
\end{tabularx}
\label{tab:Quantitative comparison}
\end{table}

\begin{figure}[!b]
    \vspace{-0.2cm}
      \centering
      \includegraphics[scale=0.27]{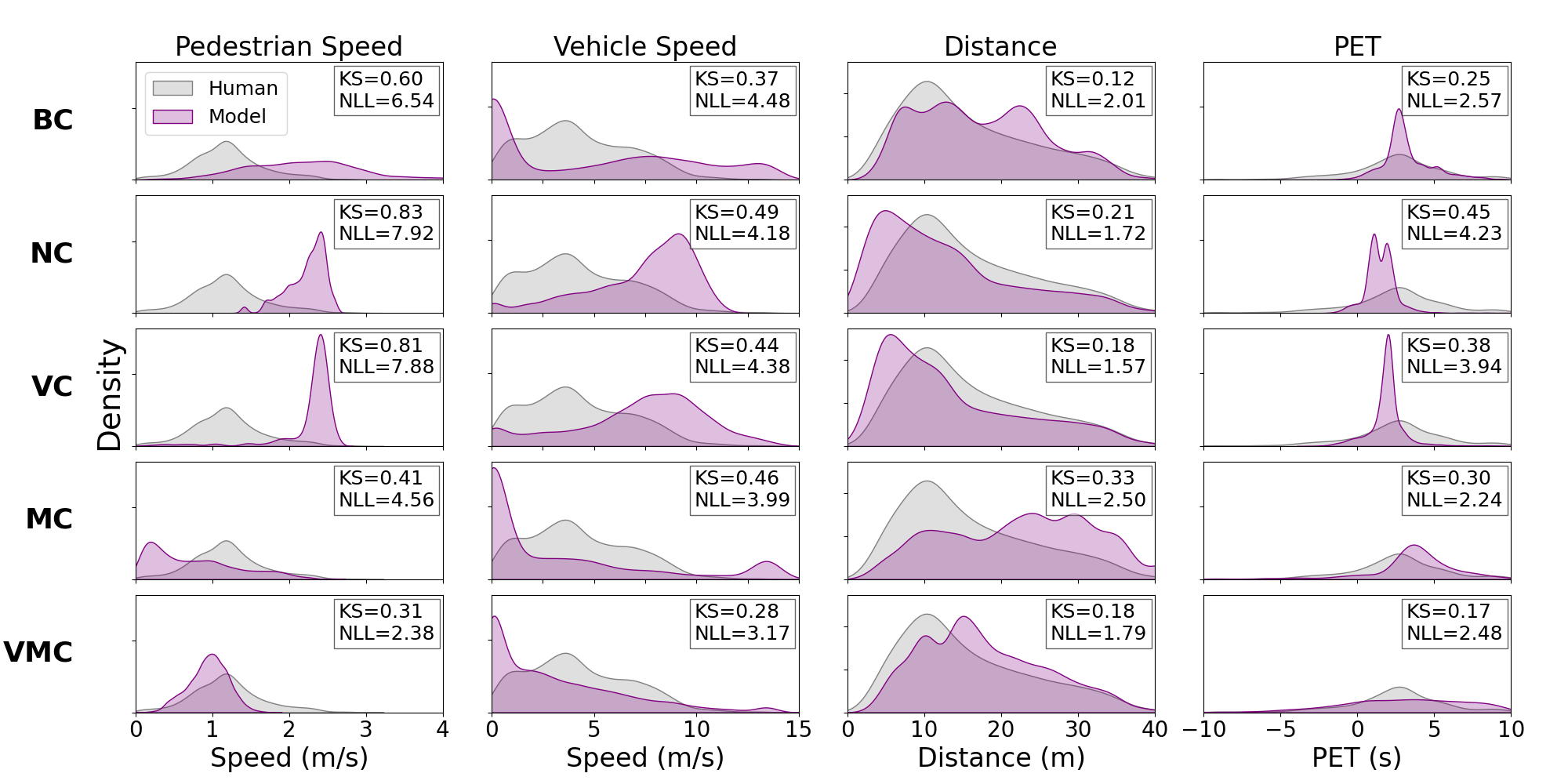}
       \vspace{-0.3cm}
      \caption{Comparison of model and real-world behavioural distributions across all time steps and trajectories for four key interaction metrics: pedestrian speed, vehicle speed, inter-agent distance, and projected post-encroachment time (PET). Each row corresponds to a different model variant. The Kolmogorov-Smirnov (KS) statistic shown in each subplot quantifies the similarity between model and real-world distributions, with lower values indicating closer alignment.}
      \label{fig:Distribution_plot}
      \vspace{-0.2cm}
\end{figure}

\subsection{Comparing model and real-world behaviour metric distributions}
\label{subsec:Distribution results}

To assess distributional similarity between model behaviours and real-world data, we report per-metric NLLs ~\citep{montali2023waymo}, together with KS statistics, as introduced in Section~\ref{subsec:Fitting of non-policy parameters}.

As shown in Table~\ref{tab:Quantitative comparison}, the VMC model achieved the lowest overall NLL (2.42), indicating the best overall alignment with real-world trajectories. This was followed by the MC and VC models, while the NC and BC models showed relatively higher NLL scores, reflecting poorer fit to the real-world data.

To provide more detail, \figurename~\ref{fig:Distribution_plot} shows per-feature comparisons for four representative metrics: pedestrian speed, vehicle speed, inter-agent distance, and projected PET. For each feature, we report the NLL together with the KS statistic as complementary measures. Positional features such as $x$ and $y$ coordinates are not shown in \figurename~\ref{fig:Distribution_plot}; however see the trajectory-level comparisons in the next section
Additional behavioural metrics used in fitting, are presented in Appendix A (\figurename\ref{fig:Distribution_plot_appendix}).

Across the four plotted metrics, the VMC model provided the closest match to real-world data, showing the lowest NLL and KS values across all metrics, as illustrated in \figurename~\ref{fig:Distribution_plot}. Although the MC model attained a slightly lower NLL for PET, (2.24 vs.\ 2.48), the VMC still achieved the lowest KS, indicating the best overall distributional similarity.

In contrast, the NC and VC models yielded much poorer fits for pedestrian speed, i.e., high per-feature NLLs (7.92 and 7.88) and large KS values (0.83 and 0.81), likely because they lack a walking-effort cost, as shown in the second and third rows of the first column in \figurename~\ref{fig:Distribution_plot}. 
The BC model aligned relatively well in distance (KS = 0.12, NLL = 2.01) and PET (KS = 0.25, NLL = 2.57), but failed to reproduce the real-world pedestrian-speed distribution.

\subsection{Comparing model and real-world trajectories}
\label{subsec:Trajectory results}

We next evaluated the spatial and temporal similarity between model and real-world trajectories using average displacement error (ADE) and final displacement error (FDE). ADE measures the average point-wise distance between model and real-world trajectories over time, while FDE quantifies the distance at the final timestep. Lower values indicate better alignment with real-world human behaviour. Summary results are shown in Table~\ref{tab:Quantitative comparison}, with full trajectory visualisations provided in Appendix A.

\figurename~\ref{fig:Trajectory_plot} presents three representative interactions: a vehicle-first case (interaction 18), a pedestrian-first case with vehicle yielding (interaction 33), and a pedestrian-first case without vehicle yielding (interaction 67). In this figure, the BC model is deterministic, so the model trajectories are identical and appear as a single purple line. In the NC model, there is no visual noise and the only stochasticity comes from the policy network, which is small, so the trajectories look as if they have merged into one. As schematically shown in the rightmost panel, in this type of plot, if the trajectory passes above the origin, this indicates that the pedestrian crossed before the vehicle, and if the trajectory passes below the origin, the vehicle went first. In \figurename~\ref{fig:Trajectory_plot}, model performance can be assessed by how close the model trajectories are to the real-world trajectory, with closer alignment indicating higher fidelity. Good agreement can also be reflected by the model position at 3~s (purple dot) coinciding with the human position at 3~s (black dot).

\begin{figure}[!t]
    \vspace{-0.2cm}
      \centering
      \includegraphics[scale=0.38]{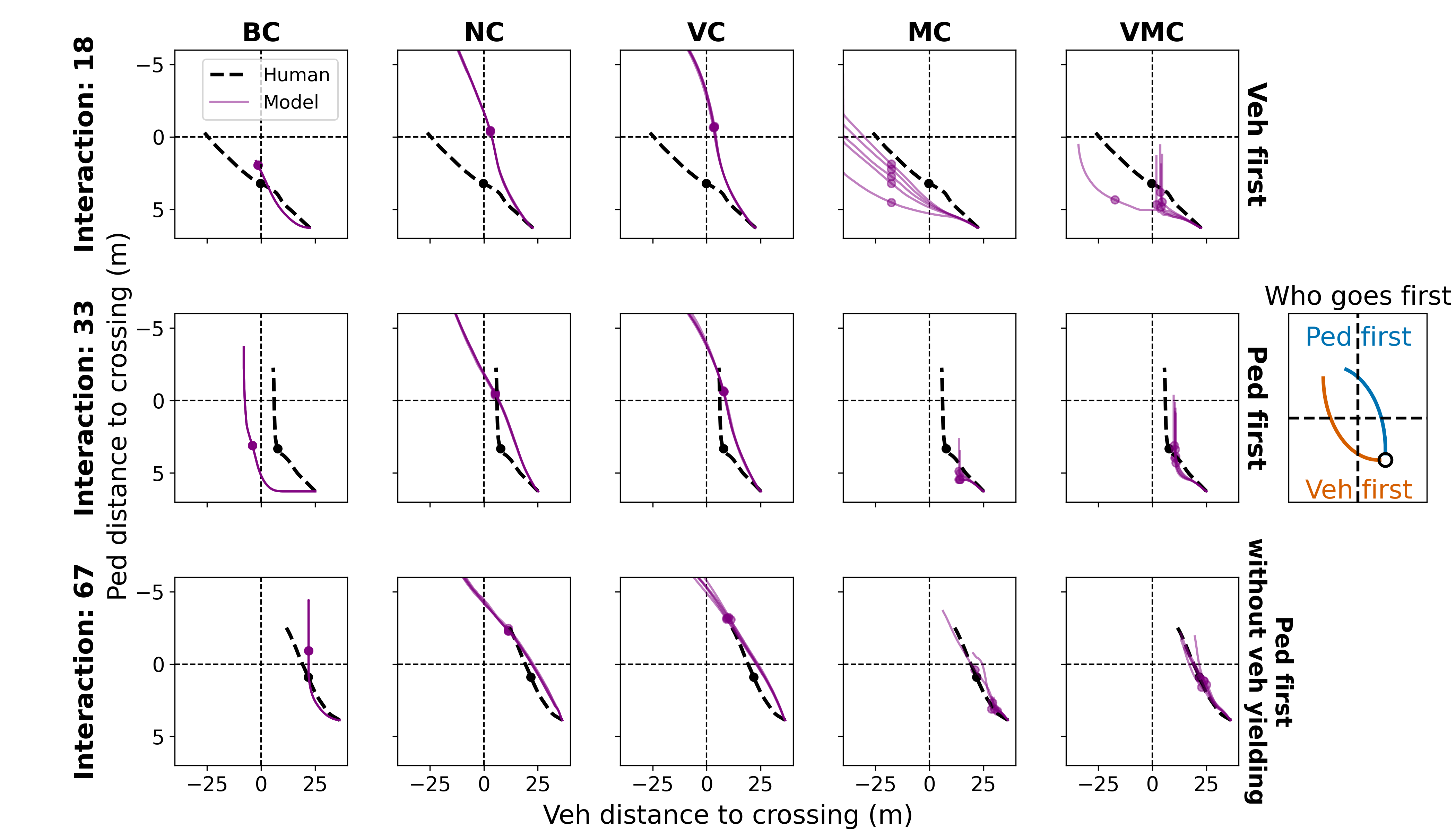}
      \vspace{-0.3cm}
      \caption{Comparison between model and real-world trajectories for three representative pedestrian-vehicle interactions (IDs 18, 33, and 67). Model trajectories are shown in purple. The $x$-axis represents the vehicle’s longitudinal distance to the crossing point; the $y$-axis shows the pedestrian’s lateral distance. Purple and black dots mark the 3-second positions of the model and real-world trajectories, respectively. Each row corresponds to a different interaction type: vehicle-first (ID 18), pedestrian-first with vehicle yielding (ID 33), and pedestrian-first without vehicle yielding (ID 67). The rightmost panel provides a schematic comparison between pedestrian-first (blue) and vehicle-first (orange) scenarios, with both trajectories originating from the same starting point (black circle).}

      \label{fig:Trajectory_plot}
      \vspace{-0.2cm}
\end{figure}
In Table~\ref{tab:Quantitative comparison}, across both ADE and FDE, the VMC model achieved the lowest errors, demonstrating its ability to reproduce human-like trajectories throughout the interaction. In contrast, the BC model and RL variants lacking motor constraints (NC and VC) showed larger errors, particularly in FDE, indicating cumulative divergence from real-world trajectories.

The BC model generally performed well in vehicle-first scenarios, as shown in interaction 18. In contrast, it failed to generate human-like pedestrian-first behaviours. In interaction 33 (second row of \figurename~\ref{fig:Trajectory_plot}), the vehicle decelerated to a stop after passing the crossing, an unrealistic action not observed in real-world data, likely due to the BC model’s lack of contextual understanding of interaction. Similarly, in interaction 67, the vehicle stopped even though the pedestrian had already crossed and posed no threat, again highlighting the BC model’s limited understanding of the interaction dynamics.

Models without motor constraints (NC and VC) allowed pedestrians to cross quickly without cost, leading to the absence of vehicle-first events. As shown in the second and third columns of \figurename~\ref{fig:Trajectory_plot}, with these models, pedestrians consistently crossed first, and vehicle-yielding behaviours were not observed.

The MC model performed well in vehicle-first cases, but driver behaviour was often more assertive than in the real-world data. In interaction 18 in \figurename~\ref{fig:Trajectory_plot}, vehicles crossed with earlier and faster motion. While the MC model also captured pedestrian-first cases (rows 2 and 3), vehicle responses in yielding scenarios appeared more abrupt and occurred further away from the pedestrian compared to real-world trajectories (shown in row 2).

\begin{figure}[!t]
    \vspace{-0.2cm}
      \centering
      \includegraphics[scale=0.35]{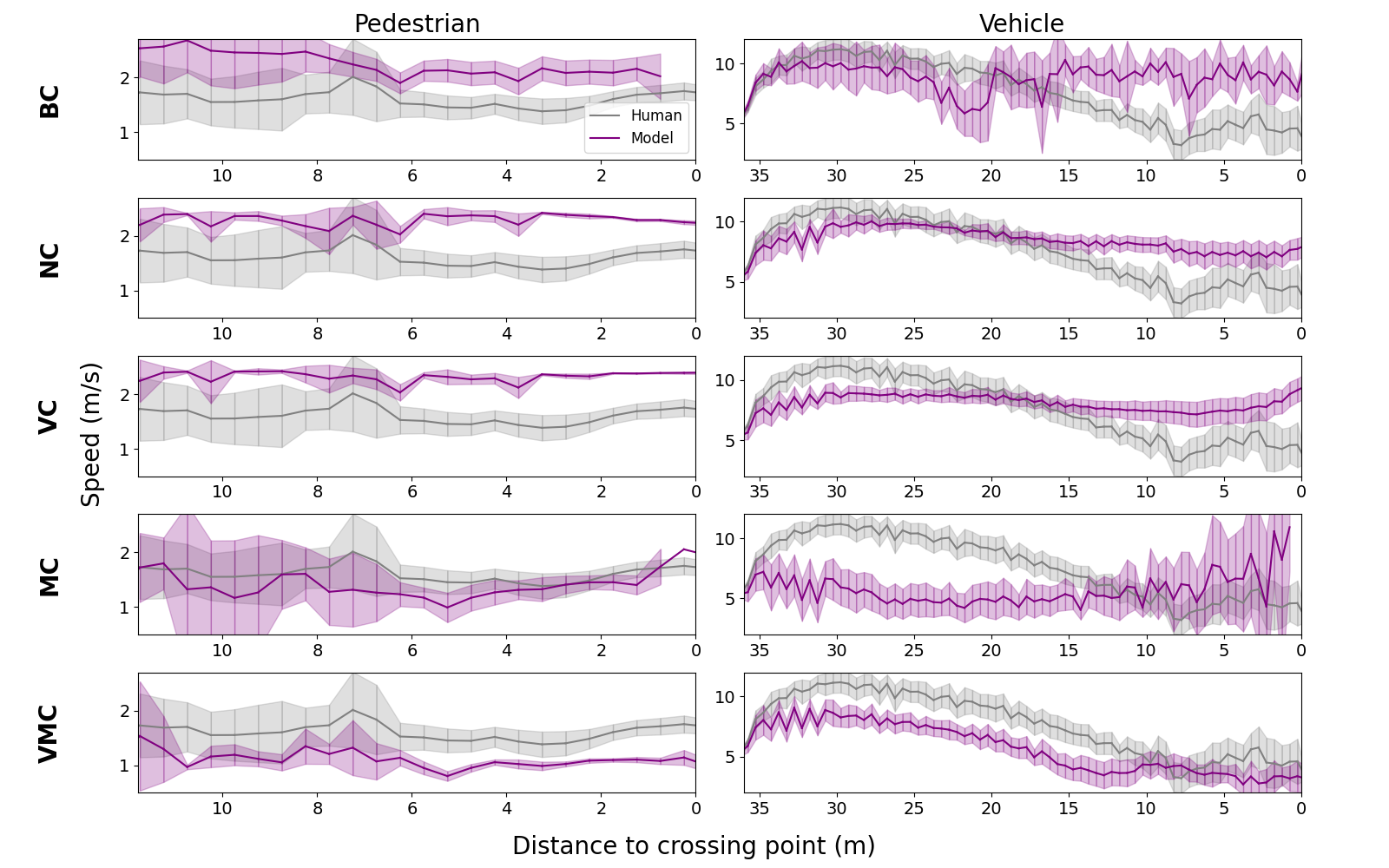}
       \vspace{-0.8cm}
      \caption{Average speed profiles of pedestrians (left) and vehicles (right). Real-world data are shown in black, and simulation results in purple. For each model, the solid line represents the mean speed, and the shaded area indicates the 95\% confidence interval. The $x$-axis represents the agent's distance to the crossing point: larger values indicate positions farther away from the crossing, while smaller values (towards the right) represent positions closer to or at the crossing. Model trajectories were first averaged across repetitions sharing the same non-policy parameters before computing group-level statistics.}
      \label{fig:Speed_distance}
      \vspace{-0.2cm}
\end{figure}

The VMC model reproduced all three interaction types observed in the real-world data, as shown in the fifth column of \figurename~\ref{fig:Trajectory_plot}. This model also captured variability within the vehicle-first category, with some vehicles yielding while others crossed first, visible in the first row of the fifth column. A strength of the VMC model is that the simulated positions at 3 s (purple dots) lie close to the real-world human positions (black dots), indicating good spatial and temporal alignment with real-world trajectories. However, a limitation is that in vehicle-first scenarios the model typically produced pedestrian-first outcomes, and when it did generate vehicle-first behaviour the vehicle showed the same aggressive driving behaviour as the MC model. It can be noted that all RL-based models were able to reproduce pedestrian-first scenarios without vehicle yielding (interaction 67), suggesting that such non-interactive patterns were relatively easier to learn regardless of constraint assumptions. However, only the VMC model showed close alignment with the real-world trajectory, as shown by the purple and black dots being close at 3 s.

\subsection{Speed of the pedestrian and driver agent}
\label{subsec:Speed result}

We further examined the average speed profiles of both the pedestrian and the vehicle as a function of their distance to the crossing point, as shown in \figurename~\ref{fig:Speed_distance}. These profiles are informative for assessing how agents adjust their motion around the crossing, for example slowing down before entry, and yielding. In \figurename~\ref{fig:Speed_distance}, both agents approached the crossing from left to right along the $x$-axis: larger values indicate positions farther from the crossing, while smaller values represent proximity to the crossing point. In the real-world data, vehicle speed gradually decreased when approaching the crossing, while pedestrians typically slowed down slightly before entering.

The BC model failed to reproduce the deceleration behaviour for vehicles. Pedestrian speed also deviated from the empirical pattern, showing only a weak reduction before crossing.

\begin{figure}[!t]
    \vspace{-0.2cm}
      \centering
      \includegraphics[scale=0.6]{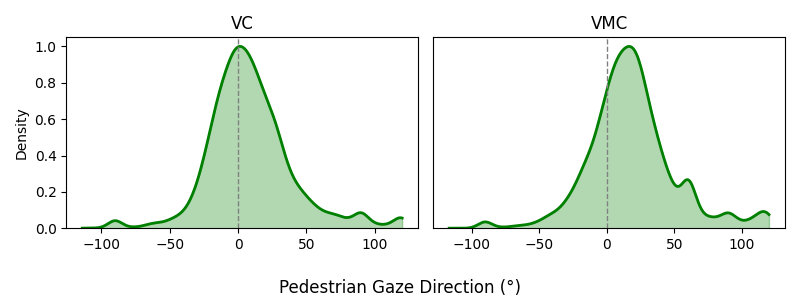}
       \vspace{-0.3cm}
      \caption{Distribution of pedestrian gaze orientation in the VC and VMC models, measured relative to the y$y$-axis (i.e., the upward direction on the map in Fig. \ref{fig:Road_layout}). A gaze direction of 0° indicates that the pedestrian is looking straight along the y$y$-axis (perpendicular to the road), while positive values indicate gaze directed toward the approaching vehicle.}
      \label{fig:Head_direction}
      \vspace{-0.2cm}
\end{figure}

In the NC and VC models, vehicle agents exhibited minimal deceleration, while pedestrian agents maintained consistently higher speeds than observed in real-world data. This behaviour stems from the absence of motor constraints, which allowed agents to accelerate or maintain high speeds without penalty. As a result, both models failed to replicate the typical slowing-down pattern seen in real-world interactions.

In contrast, the MC and VMC models more accurately captured the deceleration patterns of both agents. Pedestrian agents in these models slowed down before crossing (around $x$ = 6 m) and accelerated afterwards, aligning more closely with real-world human behaviour. Vehicle agents in the MC model applied stronger and earlier braking than observed in the real-world data, followed by a more abrupt acceleration, as shown in the second column of the fourth row of \figurename~\ref{fig:Speed_distance}. The VMC model better reproduced the observed trends, although it still exhibited slightly earlier braking than human drivers (around $x$ = 15 m) and both agents maintained overall lower speeds compared to the real-world data, as illustrated in the second column of the fifth row.

\subsection{Gaze Direction under Visual Constraints}
\label{subsec:Head Direction result}

\begin{figure}[!t]
    \vspace{-0.2cm}
      \centering
      \includegraphics[scale=0.36]{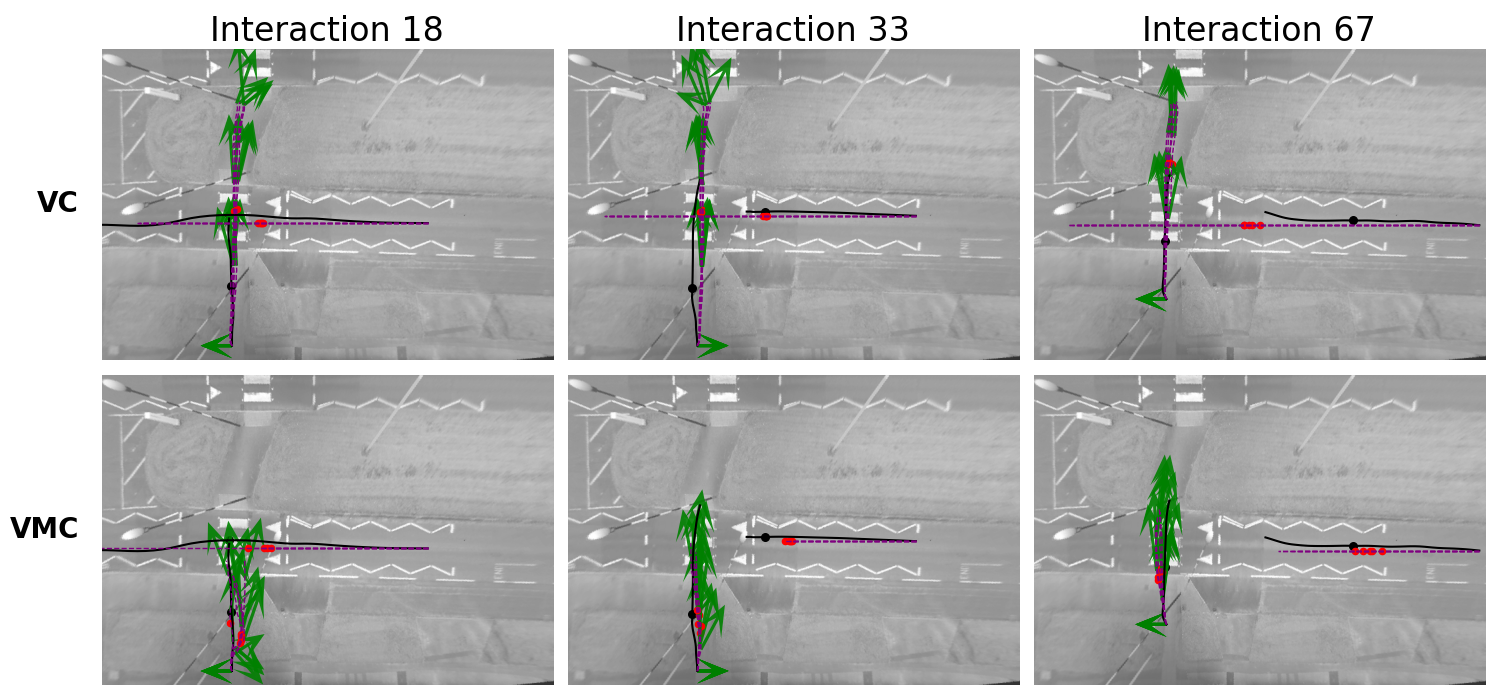}
       \vspace{-0.3cm}
      \caption{Comparison between model and real-world trajectories for the VC and VMC models across three representative interactions. Real-world trajectories are shown in black, and model trajectories in purple. Green arrows indicate the pedestrian's gaze orientation at 2 s intervals. Red and black dots denote the position of the pedestrian after 3 seconds in the model and real-world data, respectively. The background image corresponds to the real-world scene of the crossing. Note that the VMC model shows less variability than the VC model, resulting in overlapping gaze orientation arrows across repeated simulations.}
      \label{fig:Trajectory_head}
      \vspace{-0.2cm}
\end{figure}

From \figurename~\ref{fig:Head_direction}, and \ref{fig:Trajectory_head}, it can be observed that in the VMC model, pedestrians exhibited a clear tendency to direct their gaze toward the oncoming vehicle to their right. In contrast, in the VC model the gaze direction was more centrally distributed.

A closer examination of \figurename~\ref{fig:Trajectory_head} reveals more nuanced, situation-adaptive gaze patterns. For instance, in interaction 67, the pedestrian mainly looked forward, whereas in interactions 18 and 33, the gaze shifted rightward early in the interaction when the vehicle approached, and some leftward looking late in interaction 18. These patterns suggest that the VMC pedestrian learned to monitor the vehicle when it was relevant to crossing safety, leading to a situation-adaptive gaze distribution.

This behavioural difference likely arises from the reward structure of the model. The VMC model includes motor constraints, which means the pedestrian cannot cross the road fast without penalty. As a result, the pedestrian must carefully monitor the vehicle to cross safely and avoid the collision penalty. In contrast, the VC model lacks such constraints: pedestrians can cross quickly and are rarely affected by the vehicle. As discussed in Section~\ref{subsec:Trajectory results}, the VC model failed to reproduce vehicle-first or vehicle-yielding scenarios, reducing the need for vehicle monitoring. Since gaze direction only matters when avoiding collisions, the learned policy may default to centre-looking or random gaze movements when vehicle monitoring is unnecessary.

\section{Discussion}
\label{sec:Discussion}
\subsection{Main findings}
\label{subsec:Main Findings}

In this work, we present a novel multi-agent RL framework that explicitly incorporates human-like perceptual and motor constraints. Some prior studies have explored partial observability in RL-based road user agents. For example, the multi-agent RL framework by \citet{vinitsky2022nocturne} included a restricted field of view, and \citet{cornelisse2025building} leveraged partial observability to improve agent robustness for AV benchmarking. However, these works did not incorporate human-inspired elements such as distance-dependent visual noise or biomechanically constrained motor control, as included in our model to better reflect human constraints when interacting with the real world.

To make the most effective use of the limited real-world data, all models were trained and evaluated on the full dataset. Given that the RL models contain relatively few free parameters (non-policy parameters in this study), the risk of overfitting was considered low. Nevertheless, two potential concerns may arise from this setup. First, one might question whether the superior performance of the VMC model compared to the other RL-based models could be due to its greater flexibility arising from a larger number of non-policy parameters. Second, the training setup for the BC baseline does not follow the conventional practice in supervised learning, of separating training and validation data. To address these issues, we conducted an additional three-fold training/validation procedure across all 2 s trajectory segments. In each fold, all models were trained (supervised learning of the BC model; Bayesian fitting of non-policy parameter for the RL model variants) on two-thirds of the data and evaluated on the remaining one-third, and the models were ranked based on their average performance across these three folds.  The relative ranking of model performance using this fitting and evaluation method was exactly the same as the results in Table~\ref{tab:Quantitative comparison}, confirming that the improved performance of the VMC model was not simply a consequence of additional flexibility or the atypical  training setup for the BC baseline.

A key result is that the VMC model, which includes both visual and motor constraints, achieved the highest similarity to real-world data across behavioural, kinematic, and trajectory-based metrics, outperforming variants omitting one or both. Importantly, the VMC model also outperformed the behavioural cloning (BC) model, which was trained directly on real-world human demonstrations and contained a larger number of trainable parameters. This highlights the advantage of reward-driven learning under structured human-like constraints, particularly in data-limited settings.

Another methodological innovation of our work lies in modelling pedestrian-vehicle interactions at the population level through non-policy parameter distributions. Prior work in computational rationality modelling of human behaviour has often focused on individual-level modelling, where each agent is assigned its own parameter set based on observed behaviour \citep{chen2021adaptive,jokinen2021multitasking,wang2025pedestrian,wang2025Modeling}. While such approaches can capture inter-individual variability in controlled experiments, where each participant is observed across multiple trials, they are less suited to real-world road user modelling, where individuals are typically observed only once for a short period of time, e.g., as in our case during a single interaction event. Therefore, we instead estimated population-level distributions for motor and perceptual non-policy parameters. Our simulations are then run by drawing individual agent parameters from these fitted distributions, enabling the model to capture population-level variability rather than optimising a single parameter set for each agent.

\begin{figure}[!t]
    \vspace{-0.2cm}
      \centering
      \includegraphics[scale=0.6]{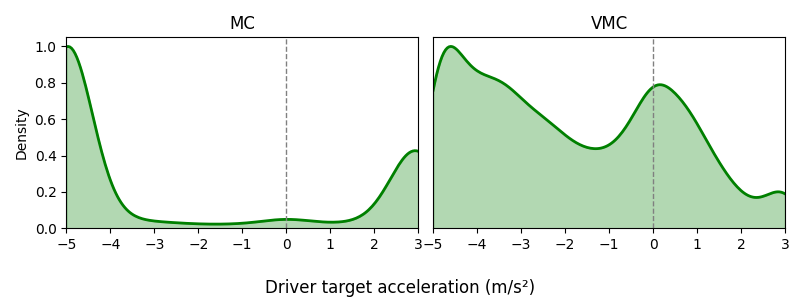}
       \vspace{-0.3cm}
      \caption{Distribution of driver target acceleration values for MC and VMC models.}
      \label{fig:Acc_plot}
      \vspace{-0.2cm}
\end{figure}

\subsection{Impact of human constraint on the pedestrian-vehicle interaction}
\label{subsec:Impact of human constraint on the pedestrian-vehicle interaction}

Here, we analyse how each type of human-inspired constraint, visual and motor, contributes to the simulated interactions. Comparing the MC and VC models indicates that introducing motor constraints (NC → MC) produces a larger improvement in interaction realism than introducing visual constraints alone (NC → VC). Although both models achieve similar NLL scores (Table~\ref{tab:Quantitative comparison}), only the MC model reproduces key behavioural patterns such as vehicle-first and vehicle-yielding scenarios. This reflects the role of motor effort in regulating movement timing: by penalising abrupt acceleration or fast crossings, motor constraints lead to smoother and more human-like behaviours. In contrast, the VC model allows the pedestrian to cross rapidly without cost, which reduces the need for the adaptation to the vehicle behaviour. As a result, interaction patterns observed in real-world data are not reproduced.

Adding visual constraints on top of motor ones (i.e., MC → VMC) further improves realism, particularly in driver responses. As shown in \figurename~\ref{fig:Acc_plot}, the MC vehicle agent often brakes early and sharply, potentially due to overconfidence in its predictions about pedestrian behaviour, to avoid the not yielding penalty. In the VMC model, perceptual uncertainty introduced by visual noise encourages more cautious deceleration, consistent with the human driver deceleration profiles shown in \figurename~\ref{fig:Speed_distance}.

Together, these findings suggest that both visual and motor constraints are necessary for realistic pedestrian-vehicle interaction modelling, but in different ways. Motor constraints primarily shape the timing and smoothness of actions, while visual constraints enhance behavioural realism by inducing more hesitant responses and better capturing the variability observed in human decision-making under uncertainty. However, it should be noted that our ablations are relatively coarse: the MC variant bundles pedestrian ballistic step execution and effort cost with driver acceleration filtering, and the VC variant bundles distance-dependent retinal noise, Bayesian estimation, and gaze-dependent acuity. Below, we discuss further opportunities to build on our work.


\subsection{Implications and future work}
\label{subsec:Implications and future work}

The current model extends our earlier framework to a multi-agent setting using real-world pedestrian-vehicle interaction data. This allows both agents to adaptively respond to each other in more realistic scenarios, bringing the framework closer to practical AV applications.

For example, with regard to AV testing, realistic human behaviour models can support the development of socially compatible AV planners, by providing training environments where AV policies learn to interact with pedestrians in ways that align with human expectations. In our case, the proposed framework enhances realism in two ways. First, its multi-agent reinforcement learning formulation enables agents to adaptively co-evolve during training, rather than only reproducing pre-specified trajectories. Second, the VMC variant not only reproduces realistic motion patterns but also simulates gaze orientation as a proxy for head movement behaviour, thereby enabling analysis of how AV systems might interpret pedestrian intent from observable cues such as head pose and facing direction. This aspect connects directly to research on pedestrian intent recognition, where algorithms commonly use head orientation to infer crossing intention or attention state \citep{rasouli2017understanding,perdana2021early}. While several empirical studies have examined pedestrians’ head and gaze orientation prior to crossing and their relation to crossing decisions \citep{zhao2023pedestrian,theisen2025head}, few modelling approaches have explicitly coupled gaze direction with the full crossing process. The present framework therefore contributes toward bridging this gap by endogenously generating gaze behaviour as part of the decision-making policy. Nonetheless, the current account of visual attention remains simplified. Future work could incorporate mechanisms such as noisy perception of both the pedestrian’s goal and surrounding agents, to capture attention sharing between task goals and other road users, as well as effort-related costs associated with head or eye movements \citep{koevoet2025effort}.

Methodologically, our ablation analysis should be interpreted at the mechanism-family level rather than as single-component attributions, since each variant combines multiple perceptual or motor elements. Future work could include finer ablations, for example separating pedestrian-only and driver-only motor constraints and disentangling visual subcomponents such as noise-only, Kalman-only, and gaze-acuity-only.

In addition to such ablation-focused refinements, several aspects of the environment design could also be improved. The current driver model lacks explicit penalties for abrupt acceleration or deceleration, which in our simulations sometimes led to unrealistic velocity fluctuations, particularly visible at the beginning of the vehicle speed profiles (left side of the right-hand panels in \figurename~\ref{fig:Speed_distance}). Future work could incorporate explicit smoothness-related rewards, such as penalties on acceleration or jerk, to better reflect the biomechanical and comfort constraints observed in human driving behaviour \citep{todorov2002optimal}. Moreover, our simulated environment did not include a representation of the fact that the real-world zebra crossing was raised, functioning as a `speed bump' for the driver. We experimented with adding a speed bump as a contextual factor intended to elicit braking, but our model drivers in the simulation either ignored it or stopped before it, and the modification did not provide further improvements in model human-likeness.

A further limitation is that we manually fixed several non-policy parameters, such as the time cost coefficient, rather than estimating them from data. We experimented with including these as additional free parameters in the modelling pipeline, but this substantially increased training time and reduced convergence stability. These results suggest that the specific RL approach we used here may struggle to handle more complex RL environments, such as those including contextual features like speed bumps or variable time costs, without additional architectural or algorithmic advances. Future work could address this by exploring more advanced policy representations, for example recurrent or attention-based neural networks that can better encode history and context, or by applying more explicit forms of inverse reinforcement learning (IRL), going beyond the parameter fitting used here to directly infer reward functions from human demonstrations. In addition, in this study, we used the SAC algorithm rather than the Proximal Policy Optimisation (PPO) algorithm \citep{schulman2017proximal}. Initial experiments using PPO revealed that the agent frequently selected boundary values in the action space. While the primary focus of this work was on RL environment design and human-like constraints, future research should also consider the role of RL algorithm choice and optimisation in shaping policy behaviour. The selection of RL algorithms may also affect the stability, and realism of learned behaviours.

Another aspect relating to evaluation robustness concerns the BC baseline. As noted above, we trained and evaluated BC on the entire dataset to align conditions with the RL setup in a data-limited setting. This supports high one-step, open-loop accuracy on the same data, but closed-loop rollouts can accumulate errors once the policy drifts away from the demonstrated trajectory. Future work could address this limitation by training the BC model on a larger and more diverse dataset to better examine its behaviour across varied pedestrian-vehicle interactions.

We argue that the superior performance of the VMC model, relative to the other model variants, primarily reflects structural changes introduced by the perceptual and motor constraints rather than the larger number of non-policy parameters. This interpretation is supported by the training/validation split analysis we mentioned above, where the relative performance ranking remained consistent across folds, indicating that the improved performance of the VMC model is unlikely to result from overfitting due to additional parameters. However, we did not test this using metrics such as the Akaike Information Criterion (AIC), which balances model fit and complexity by penalising the number of free parameters \citep{akaike2003new}. Our present evaluation followed \citet{montali2023waymo} in relying on a composite NLL, whereas AIC requires likelihoods computed at the level of individual observations. Future work could therefore either develop a way to compute AIC from our setting, for example by deriving standardised individual-level likelihoods, or explore alternative criteria that explicitly balance goodness of fit against model complexity. Such approaches would help distinguish improvements due to human constraints from those arising from additional flexibility.

Finally, this study focused on simplified one-to-one interactions between a single pedestrian and vehicle. While this setup supports simpler training and interpretable analysis, extending the framework to model groups of pedestrians, and dense traffic environments remains an important next step for real-world scalability and generalisation.

\section{Conclusions}
\label{sec:Conclusions}

This study developed a multi-agent reinforcement learning framework that integrates human-like sensory and motor constraints to model pedestrian-vehicle interactions at unsignalised crossings. We evaluated four model variants—No Constraint (NC), Motor Constraint (MC), Visual Constraint (VC), and Visual and Motor Constraint (VMC)—on a real-world dataset using a comprehensive set of behavioural, kinematic, and trajectory-based metrics. The VMC model achieved the highest similarity to human data across all metrics, outperforming other RL variants and a behavioural cloning (BC) model, demonstrating the value of RL-based human behaviour modelling in data-limited conditions. In addition, the VMC model provided a simple representation of gaze orientation, offering a first step toward capturing how pedestrians use gaze in interactive decision-making. By comparing models with different human constraints, we found that motor constraints result in smoother movements that reflect human-like speed adjustments during crossing interactions. The inclusion of visual constraints introduces perceptual uncertainty and field-of-view limitations, leading the agents to exhibit more cautious and variable behaviour, such as less abrupt deceleration. These findings highlight the importance of including both motor and visual constraints in interactive behaviour modelling. Another contribution of our approach lies in its use of population-level model fitting of RL-based human behaviour models. Instead of optimising a fixed parameter set for each simulated agent, we infer population-level distributions over sensory and motor constraints. This allows the model to reflect group-level behavioural tendencies while preserving individual variability. Such an approach is particularly valuable in real-world road user modelling, where repeated measurements per individual are rare, and stands in contrast to prior work primarily conducted in controlled experimental settings. The proposed framework provides a promising direction for both behavioural modelling and the development of socially acceptable AVs. Future work could extend the approach to more complex scenarios, such as group pedestrian dynamics and multi-vehicle interactions, and refine the reward design to better capture social and contextual norms in road user behaviour.



\section*{Acknowledgments}
This work was supported by the UK Engineering and Physical Sciences Research Council (grants EP/S005056/1 and EP/Z53593X/1).

\section*{Declarations of interest: none}


\newpage
\appendix
\renewcommand{\figurename}{Figure}
\renewcommand{\thefigure}{\arabic{figure}}
\section*{Appendix~A: Additional Material}
\label{sec:appendix}


\begin{figure}
    \vspace{-0.2cm}
      \centering
      \includegraphics[scale=0.20]{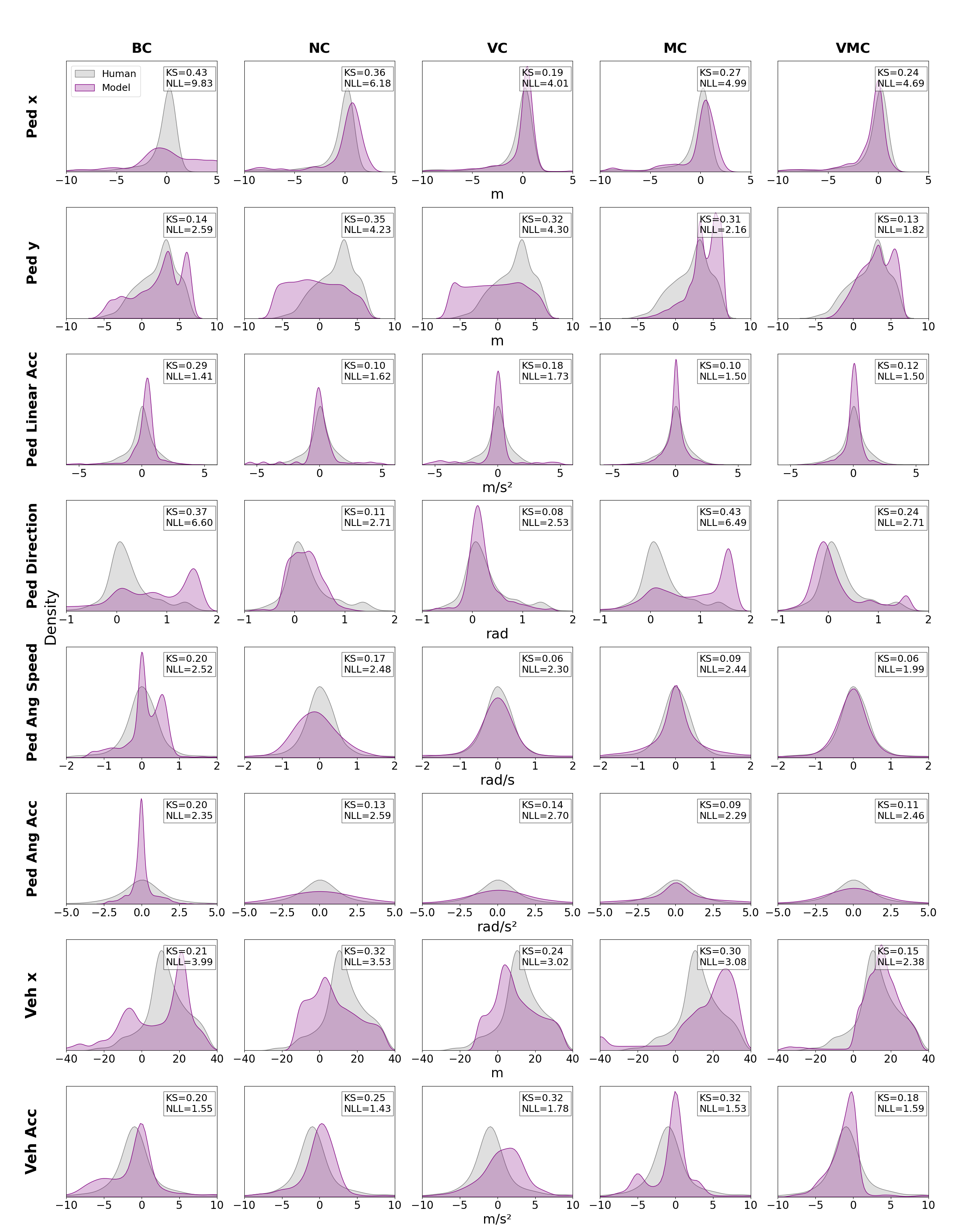}
       \vspace{-0.3cm}
       \caption{Comparison of model and real-world behavioural metric distributions across all features used in fitting (see Section~\ref{subsec:Fitting of non-policy parameters}), in addition to the four key metrics shown in \figurename~\ref{fig:Distribution_plot}. Each column corresponds to one model variant, and each row to one behavioural metric.}

      \label{fig:Distribution_plot_appendix}
      \vspace{-0.2cm}
\end{figure}


\begin{figure}[!t]
    \vspace{-1.5cm}
      \centering
      \includegraphics[scale=0.2]{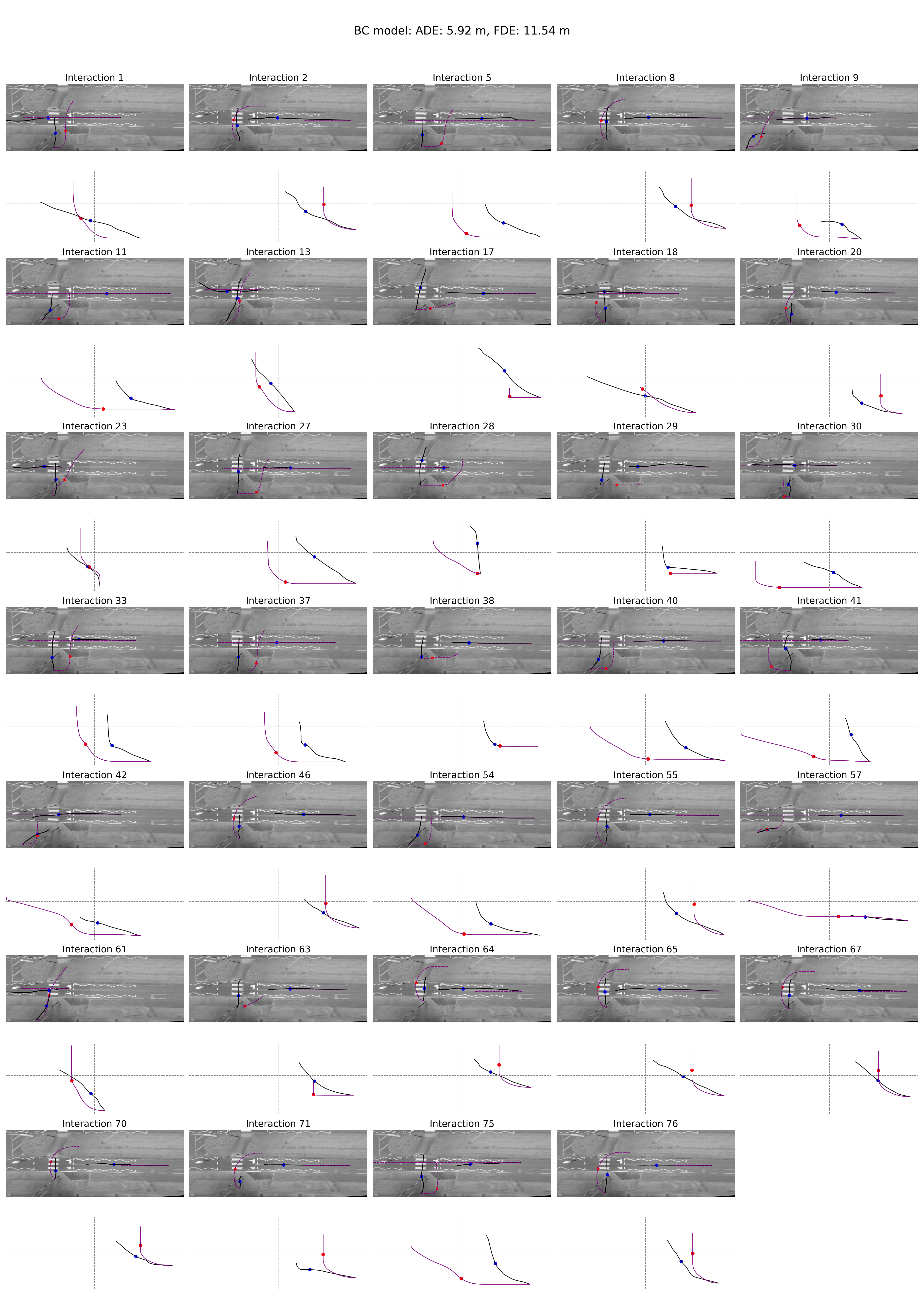}
      \caption{Model and human trajectories for the Behavioural Cloning (BC) model across all pedestrian-vehicle interactions. For each interaction, the top panel shows the trajectories overlaid on the real-world road layout, while the bottom panel depicts the relative spatial positions of the vehicle ($x$-axis: longitudinal distance to crossing point) and pedestrian ($y$-axis: lateral distance to crossing point) at each time step. Human trajectories are shown in black, and model trajectories are shown in purple. Red and black dots indicate the 3-second positions of the model and human agents, respectively.}

      \label{fig:Full_traj_BC}
      \vspace{-0.5cm}
\end{figure}

\begin{figure}[!t]
    \vspace{-1.5cm}
      \centering
      \includegraphics[scale=0.2]{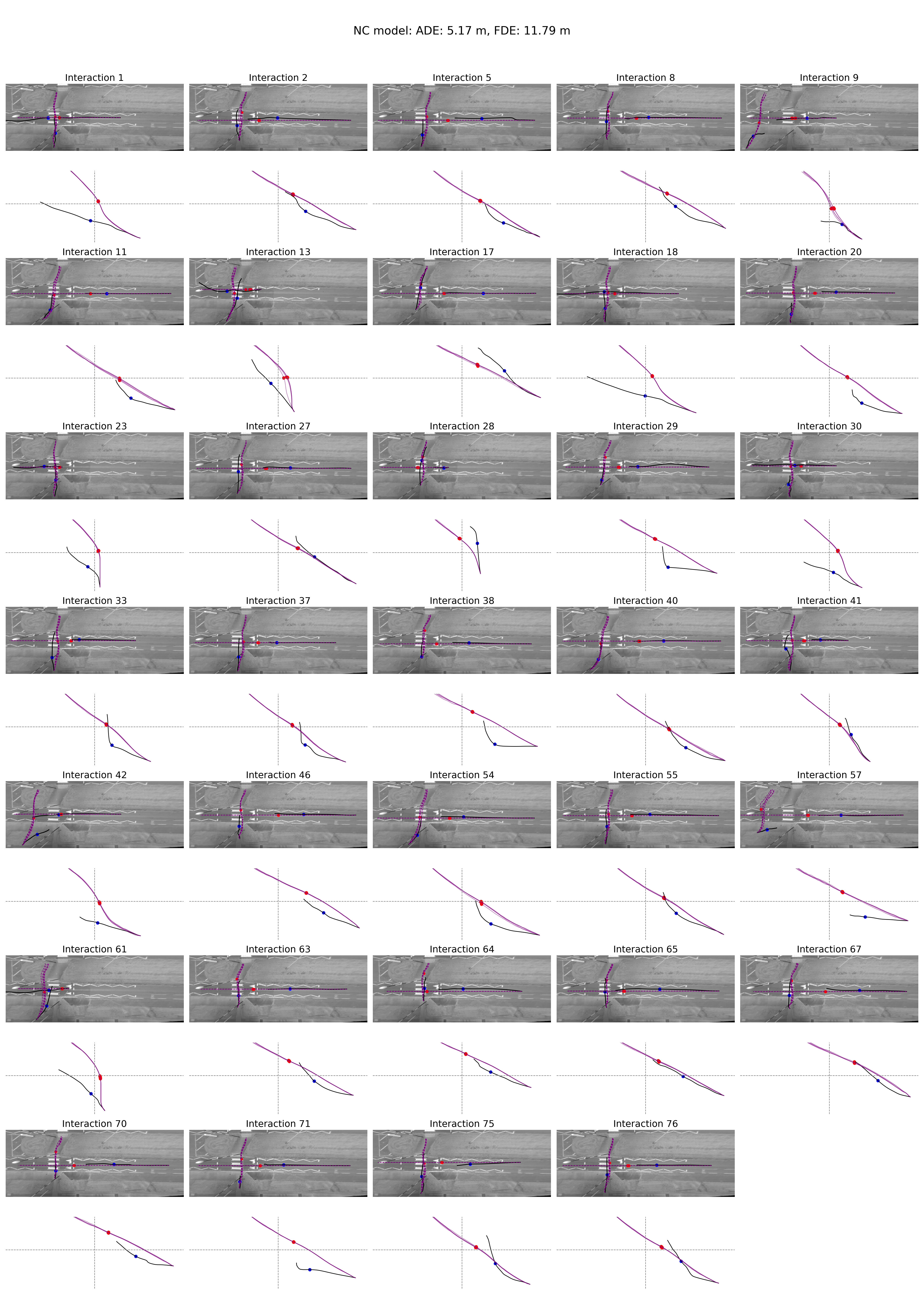}
      \caption{Model and human trajectories for the No-Constraint (NC) model across all pedestrian-vehicle interactions.}
      \label{fig:Full_traj_NC}
      \vspace{-0.5cm}
\end{figure}

\begin{figure}[!t]
    \vspace{-1.5cm}
      \centering
      \includegraphics[scale=0.2]{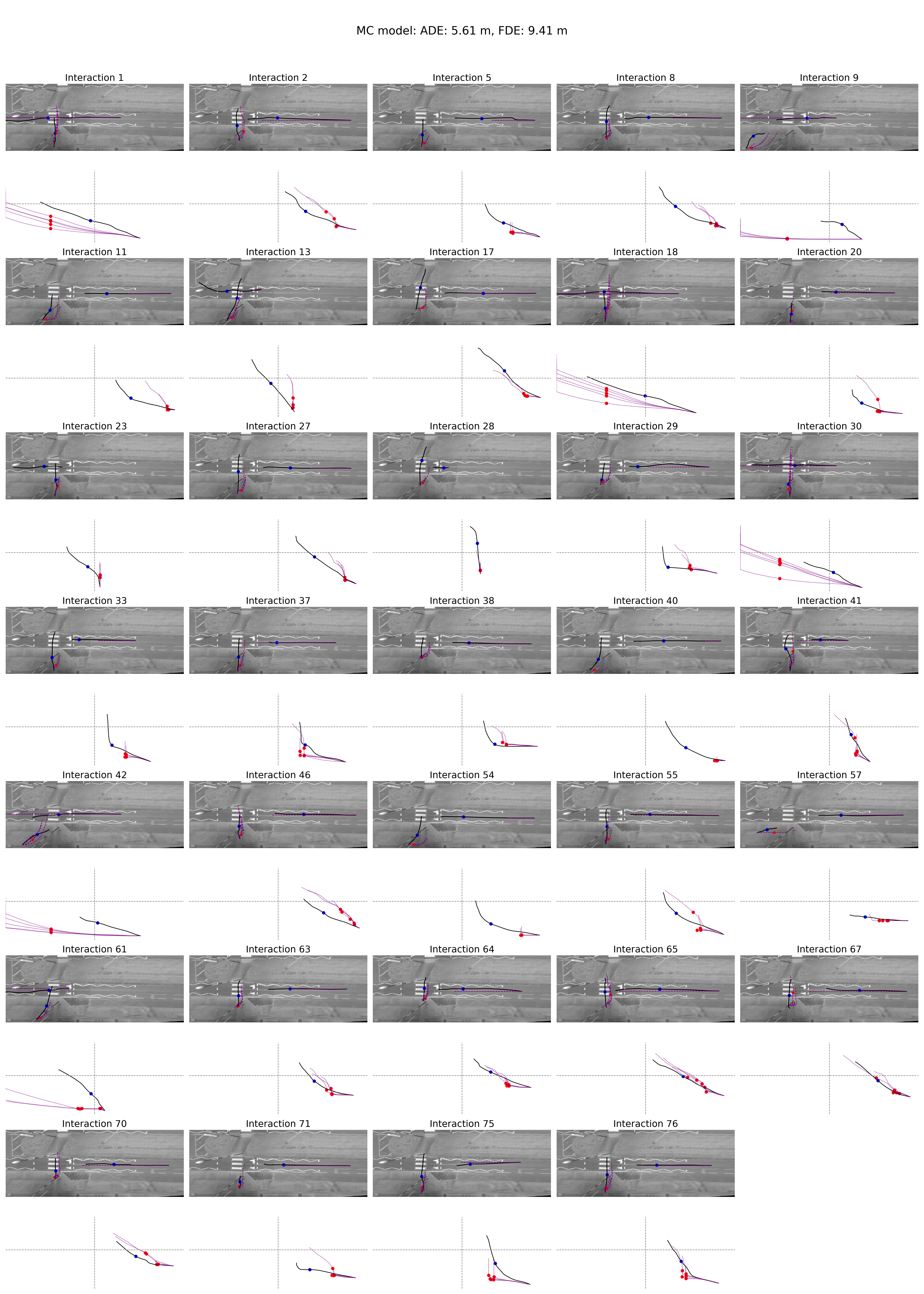}
      \caption{Model and human trajectories for the Motor-Constraint (MC) model across all pedestrian-vehicle interactions.}
      \label{fig:Full_traj_MC}
      \vspace{-0.5cm}
\end{figure}

\begin{figure}[!t]
    \vspace{-1.5cm}
      \centering
      \includegraphics[scale=0.2]{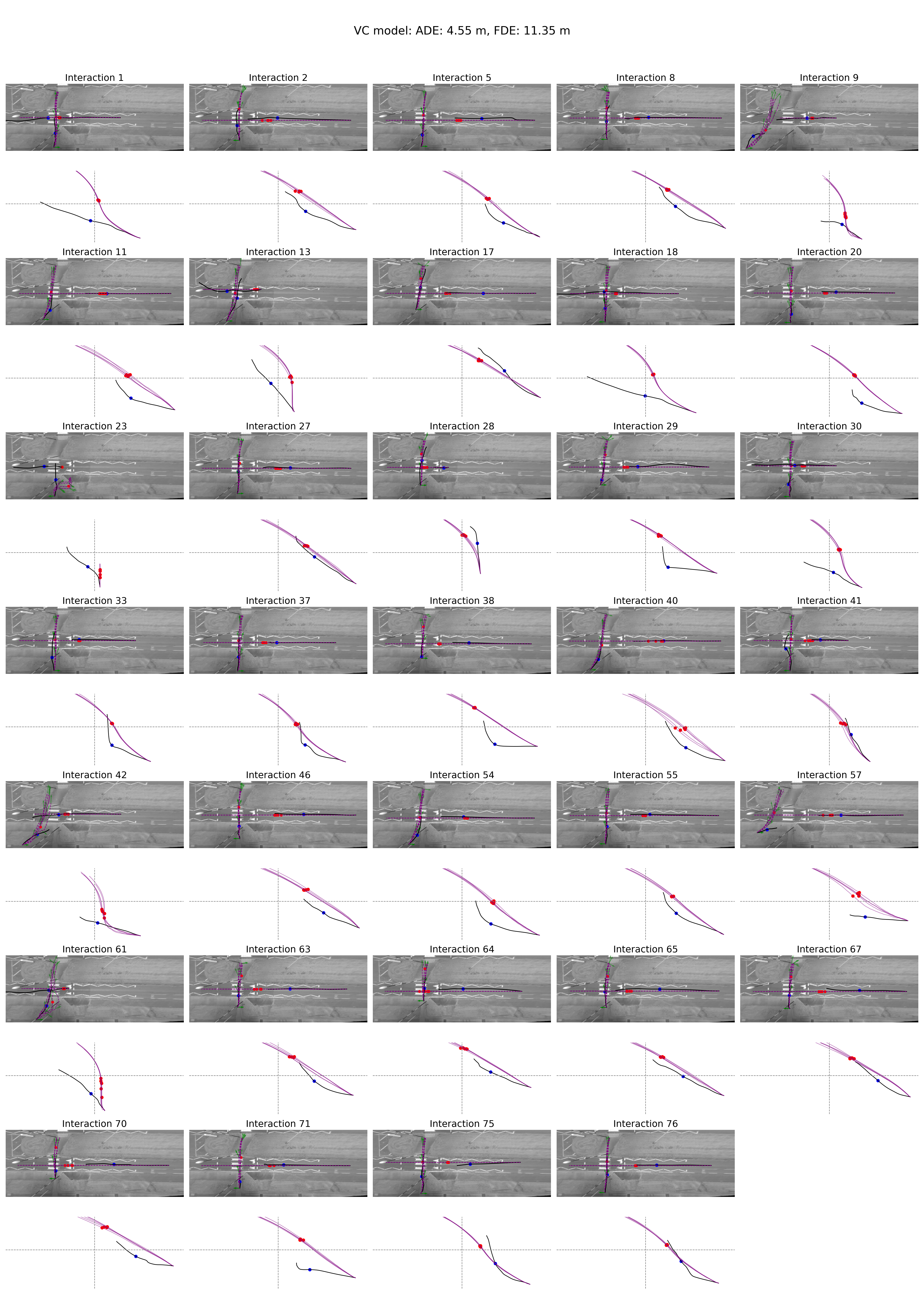}
      \caption{Model and human trajectories for the Visual-Constraint (VC) model across all pedestrian-vehicle interactions.}
      \label{fig:Full_traj_VC}
      \vspace{-0.5cm}
\end{figure}

\begin{figure}[!t]
    \vspace{-1.5cm}
      \centering
      \includegraphics[scale=0.2]{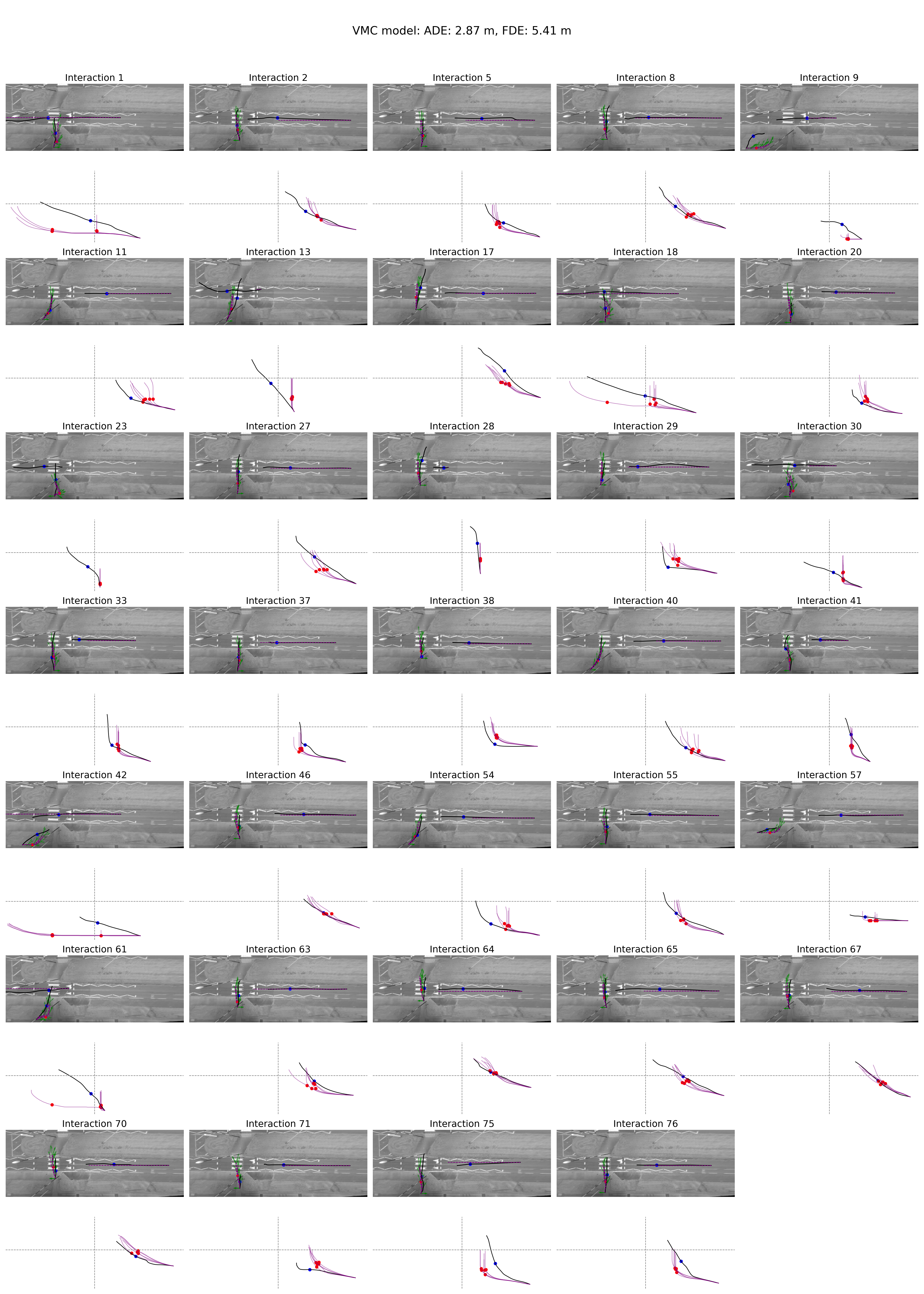}
      \caption{Model and human trajectories for the Visual and Motor-Constraint (VMC) model across all pedestrian-vehicle interactions.}
      \label{fig:Full_traj_VMC}
      \vspace{-0.5cm}
\end{figure}

\clearpage
\bibliographystyle{elsarticle-harv} 
\bibliography{cas-refs}





\end{document}